# Towards Intelligent Geospatial Data Discovery: a knowledge graph-driven multi-agent framework powered by large language models

Ruixiang Liu[a], Zhenlong Li[a*], Ali Khosravi Kazazi[a]
[a] Geoinformation and Big Data Research Laboratory, Department of Geography, The Pennsylvania State University, University Park, PA, USA
[*]Corresponding author: zhenlong@psu.edu

**Abstract:** The rapid growth in the volume, variety, and velocity of geospatial data has created data ecosystems that are highly distributed, heterogeneous, and semantically inconsistent. Existing data catalogs, portals, and infrastructures still rely largely on keyword-based search with limited semantic support, which often fails to capture user intent and leads to weak retrieval performance. To address these challenges, this study proposes a knowledge graph-driven multi-agent framework for intelligent geospatial data discovery, powered by large language models. The framework introduces a unified geospatial metadata ontology as a semantic mediation layer to align heterogeneous metadata standards across platforms and constructs a geospatial metadata knowledge graph to explicitly model datasets and their multidimensional relationships. Building on the structured representation, the framework adopts a multi-agent collaborative architecture to perform intent parsing, knowledge graph retrieval, and answer synthesis, forming an interpretable and closed-loop discovery process from user queries to results. Results from representative use cases and performance evaluation show that the framework substantially improves intent matching accuracy, ranking quality, recall, and discovery transparency compared with traditional systems. This study advances geospatial data discovery toward a more semantic, intent-aware, and intelligent paradigm, providing a practical foundation for next-generation intelligent and autonomous spatial data infrastructures and contributing to the broader vision of Autonomous GIS.

## 1. Introduction

In the big data era, the volume, variety, and velocity of geospatial data have increased substantially (Lee & Kang, 2015; Li et al., 2016). Massive, heterogeneous, and multi-source geospatial resources have accumulated rapidly across the world, including national open data portals, regional Spatial Data Infrastructures (SDIs), and scientific data repositories (Hinz & Bill, 2021; Pampel et al., 2023). However, in contrast to this proliferation of data supply, users continue to face substantial challenges in geospatial data discovery (Al-Yadumi et al., 2021; De Andrade et al., 2014). Datasets are distributed across disparate platforms, adhere to different metadata standards, and employ inconsistent semantic representations (Wu et al., 2024). As a result, users frequently rely on keyword-based search mechanisms and iteratively refine their queries within opaque ranking systems, attempting to approximate their true information needs. This trial-and-error process makes it difficult to identify datasets that truly meet specific research or decision-making needs in an efficient and accurate manner.

Recent advances in large language models (LLMs) have started to reshape GIScience research and practice, leading to the emergence of autonomous GIS (Li & Ning, 2023), an AI-powered geographic information system that can interpret natural language, reason about spatial problems, and generate code to support automated geospatial data discovery and collection, analysis, and visualization. Geospatial data discovery includes identifying, locating, and evaluating data resources that match a user's information need through query formulation, retrieval, relevance assessment, filtering, and ranking. Because data discovery is the starting point for later spatial analysis and decision-making, its quality directly affects whether downstream analytical and modeling tasks can proceed effectively. In this context, Intelligent Geospatial Data Discovery that combines intent understanding, structured retrieval, and principled ranking should be considered as a key component of level 3 autonomous GIS (Z. Li et al., 2025), where data collection and preparation can be carried out automatically based on tasks specified by humans or agents.

To address this need, this paper proposes a knowledge graph-driven multi-agent framework for intelligent geospatial data discovery powered by large language models. The main contributions of this study are as follows.

- First, we introduce a unified semantic modeling approach for geospatial metadata that maps multiple mainstream metadata standards to a shared ontology, enabling cross-platform semantic alignment and supporting the construction of a geospatial metadata knowledge graph.
- Second, we design and implement a multi-agent collaborative workflow that integrates intent parsing, graph retrieval, and answer synthesis into an interpretable and modular pipeline, forming an end-to-end process from natural language queries to discovery results.
- Third, through systematic performance evaluation and case studies, we demonstrate that the proposed framework achieves significant improvements over traditional systems in terms of ranking performance and recall rate, while showing high intent matching accuracy and the discovery transparency.

Overall, this research advances geospatial data discovery toward a more semantic, intent-aware, and intelligent paradigm, and provides methodological foundations for the next generation of intelligent and autonomous spatial data infrastructures.

## 2. Background and related works

### 2.1 Geospatial metadata and data catalogs

Metadata is data that describes data and can be used to characterize a wide range of information resources, including datasets, services, knowledge bases, and other information objects (Danko, 2011; Haslhofer & Klas, 2010). In the geospatial domain, geospatial metadata is widely recognized as a key mechanism for organizing and managing geospatial data, as it supports the discovery, evaluation, access, and use of geospatial resources (Hu et al., 2022; Quarati et al., 2021).

To facilitate interoperability across systems and organizations, a variety of geospatial metadata standards have been proposed and widely adopted in both academic and operational communities.

The U.S. Federal Geographic Data Committee (FGDC) released the Content Standard for Digital Geospatial Metadata (CSDGM) (FGDC, 1998), which dominated geospatial metadata practices for an extended period. After that, the ISO 19115 family became mainstream standards, with ISO 19115-1:2014 serving as the core standard for describing geographic data and services (ISO, 2014). ISO 19115-2:2019 further extends this framework for an enhanced description of the acquisition and processing of geographic information, including imagery (ISO, 2019).

Moreover, several organizations and communities have proposed more lightweight or practice-oriented metadata specifications. The World Wide Web Consortium (W3C) introduced the Data Catalog Vocabulary (DCAT) as a lightweight RDF-based (Resource Description Framework) metadata model for describing datasets and data catalogs on the Web (W3C, 2014). DCAT provides a more general and flexible vocabulary designed for cross-domain data publication and interoperability in linked data environments. The SpatioTemporal Asset Catalog (STAC) emphasizes concise descriptions of spatiotemporal assets and their associated resources in cloud-based environments (Lawler et al., 2025; Simoes et al., 2021). The Open Geospatial Consortium (OGC) has established a suite of open standards that define principles for the description, discovery, and access of geospatial data. Among these, the OGC Catalogue Service for the Web (CSW) specifies metadata-based catalog service interfaces (Nebert et al., 2016). OGC API - Records adopts modern Web API design principles and is widely viewed as a key evolutionary direction beyond CSW (Vretanos et al., 2025).

Supported by these metadata standards, geospatial data catalogs have gradually evolved into core components of SDIs (Hendriks et al., 2012; Rajabifard & Williamson, 2001). Such systems are designed to centrally manage metadata resources provided by diverse data producers and to offer users a unified entry point for resource discovery, including Global Earth Observation System of Systems (GEOSS) (Boldrini et al., 2023), the Infrastructure for Spatial Information in Europe (INSPIRE) (Crompvoets et al., 2018), and the U.S. government-led open data platform Data.gov (Krishnamurthy & Awazu, 2016). Data.gov is built upon the open-source CKAN (Comprehensive Knowledge Archive Network) platform, which manages metadata using its native schema while supporting mappings to DCAT-US (Winn, 2013).

### 2.2 Geospatial data discovery

Early geospatial data discovery systems primarily relies on keyword-based retrieval mechanisms, returning candidate datasets through string matching over metadata fields such as titles, abstracts, and keywords (Bone et al., 2016; Katumba & Coetzee, 2017). These systems typically utilize general information retrieval libraries such as Apache Lucene to construct inverted indexes, enabling efficient query processing (Mai et al., 2020). However, keyword-based approaches struggle to explicitly capture spatial coverage, temporal extent, and complex semantic relationships among themes, and their effectiveness is highly dependent on metadata text quality and users' choice of terminology.

To improve retrieval accuracy and user interaction, researchers have introduced discovery mechanisms supporting multi-criteria queries through faceted filtering (Hardy & Durante, 2014). Under this paradigm, spatial extent is indexed using bounding boxes or geometric objects, temporal information is modeled as time intervals or timestamps, and semantic attributes are organized through enumerations or classification schemes (Giunchiglia et al., 2012; Hervey et al., 2020). Faceted discovery enables users to iteratively refine search results across multiple dimensions, but it still does not incorporate semantic understanding capabilities (De Andrade et al., 2014).

With the advancement of semantic web technologies and knowledge representation methods, semantic search and ontology-based geospatial data discovery have attracted increasing attention. In this line of work, domain ontologies are constructed to formally model geospatial datasets on topic, space and time, explicitly defining their hierarchical structures and semantic relationships (Klien et al., 2006; Yue et al., 2011). These approaches typically combine RDF graph models with ontology reasoning mechanisms (Gui et al., 2013; Li et al., 2014). Compared to traditional methods, semantic-driven discovery offers clear advantages in handling synonymy, polysemy, and hierarchical concept relations (Bogdanović et al., 2015).

With the emergence of LLMs, the research paradigm of geospatial data discovery is undergoing a fundamental transformation. LLM-Find adopts a plug-and-play design that automates data retrieval by developing agent skills using source-specific handbooks (Ning et al., 2025). GDQA integrates a geospatial data knowledge graph with LLMs to navigate and infer over structured graph representations for geospatial data discovery (H. Li et al., 2025). Despite these promising developments, existing efforts remain at an early stage, which either focus on retrieval within specific data sources or provide preliminary explorations without establishing a scalable and unified framework. This paper aims to achieve intelligent geospatial data discovery within the autonomous GIS paradigm and provide a systematic foundation for future research in this emerging direction.

## 3. Methodology

This study proposes a knowledge graph-driven, multi-agent framework powered by LLM, which conceptualizes geospatial data discovery as an intelligent, multi-stage collaborative reasoning process (Fig. 1). The framework consists of three integrated components: a system-level orchestration module, a multi-agent discovery pipeline, and a geospatial knowledge graph. Through unified workflow control and state and memory management, the framework supports natural language-driven data discovery tasks. Each agent follows the perception-reasoning-action cognitive cycle, which can dynamically interpret environmental inputs, perform semantic and algorithmic reasoning, and execute appropriate actions through external tool invocation and interaction with modules.

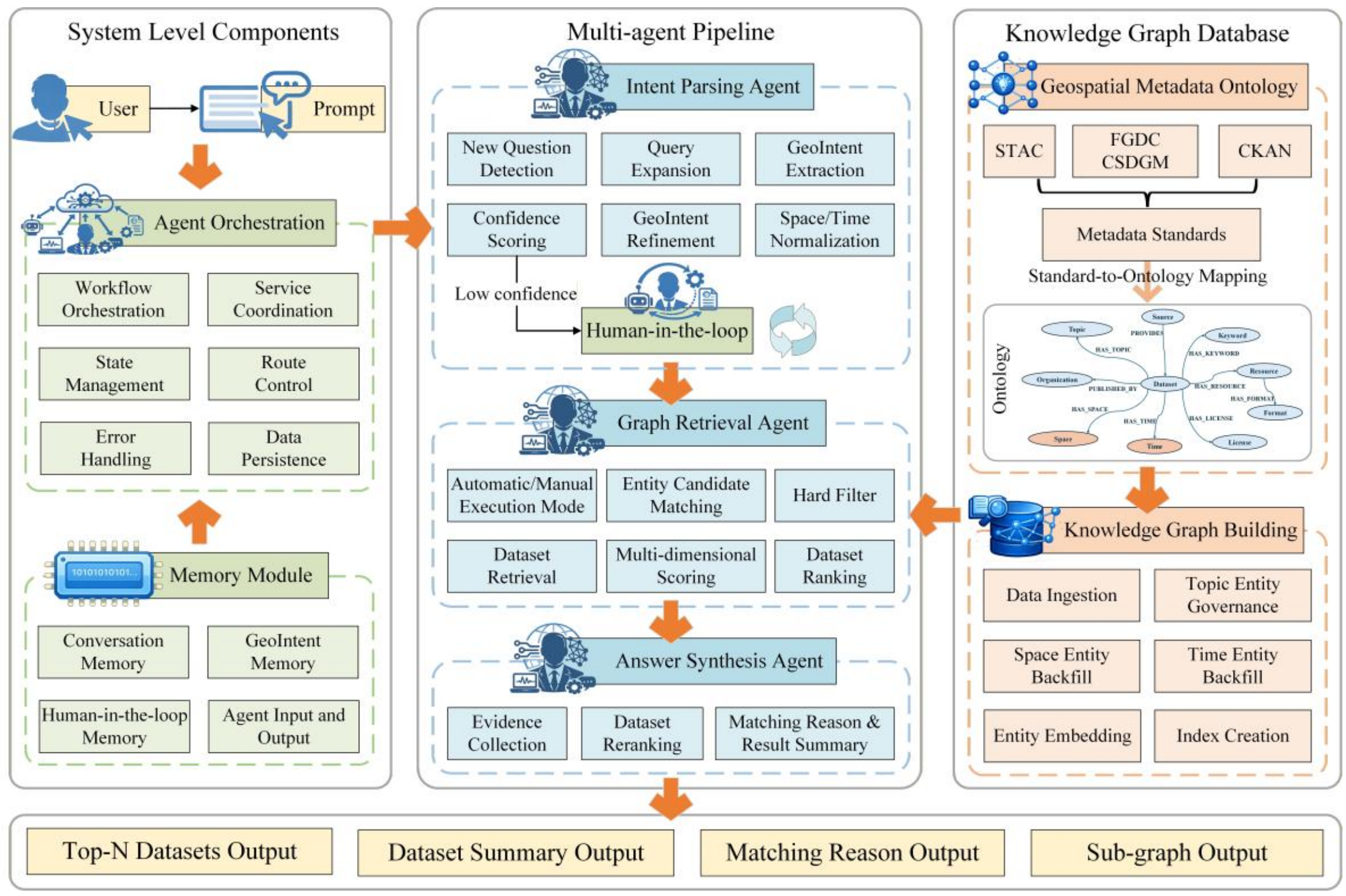

Fig. 1. Overall architecture of the intelligent geospatial data discovery framework.

## 3.1 Geospatial metadata knowledge graph

### 3.1.1 Metadata ontology design

To support intelligent and intent-aware geospatial data discovery, this study designs a geospatial metadata ontology that enables unified modeling and semantic alignment of heterogeneous metadata descriptions derived from different metadata standards. Rather than attempting to replace existing metadata standards, the proposed ontology functions as a semantic mediation layer and provides a consistent semantic foundation for higher-level intelligent discovery tasks.

The ontology design follows two principles: standard compatibility and semantic abstraction with minimal sufficiency. First, the ontology ensures that existing metadata from different catalogs can be mapped without semantic loss. Second, only concepts and relationships that are essential for data discovery and understanding are retained, avoiding the direct replication of complex and highly redundant structures present in individual standards.

The ontology is informed by and integrates several widely used geospatial metadata standards, including STAC, FGDC CSDGM, and CKAN. Table. A1 presents the mapping of the proposed ontology and the three standards. Fig. 2 shows the logical model of the geospatial metadata ontology. At the conceptual level, the geospatial metadata ontology adopts *Dataset* as the core entity, representing geospatial datasets that can be discovered and accessed. *Topic* and *Keyword* are used to represent the thematic semantics and content labels of a dataset, corresponding respectively to higher-level conceptual themes and finer-grained keyword descriptions. *Space* and *Time* characterize the spatial and temporal coverage of the dataset, supporting data discovery under spatial and temporal constraints. *License* describes the usage and sharing restrictions associated with the dataset. *Organization* represents the entity responsible for publishing or maintaining the dataset, while *Source* denotes the catalog, platform, or system from which the dataset originates. The *Resource* entity further captures concrete access resources or service endpoints associated with a dataset, such as download links or service interfaces, and *Format* specifies the actual data or service format through which a *Resource* is made available.

At the relational level, the ontology employs explicit semantic relationships to represent structured connections among entities, including *HAS_TOPIC*, *HAS_KEYWORD*, *HAS_SPACE*, *HAS_TIME*, *HAS_FORMAT*, *HAS_LICENSE*, *HAS_RESOURCE*, *PUBLISHED_BY*, and *PROVIDES*. Each relationship is defined with a clear semantic orientation, characterizing the dependency and association between a *Dataset* and its descriptive entities.

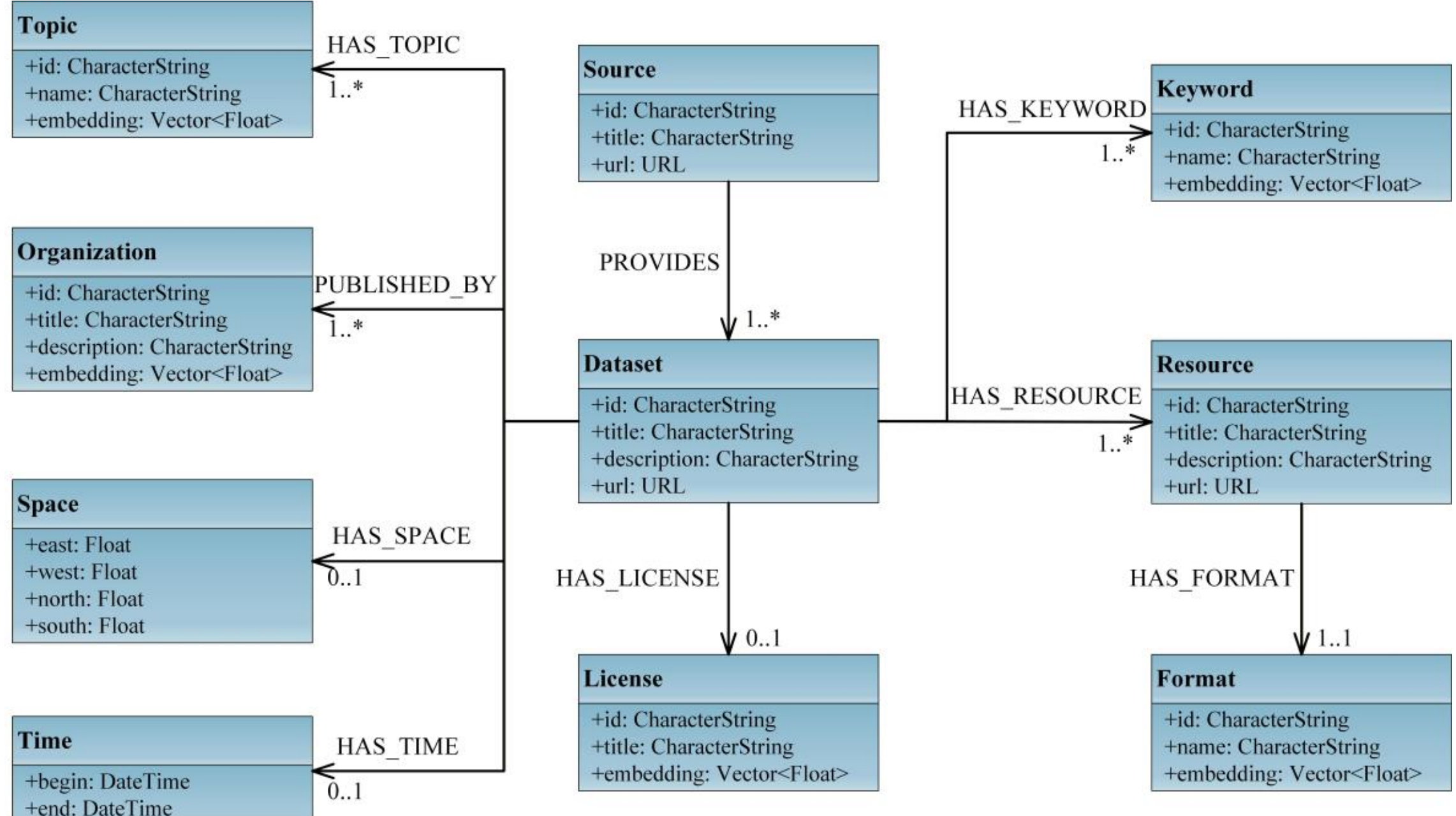

Fig. 2. Logical model of the geospatial metadata ontology.

### 3.1.2 Knowledge graph building

The knowledge graph building follows a multi-stage pipeline architecture that progressively transforms raw metadata from STAC, FGDC CSDGM, and CKAN into a structured, semantically consistent, and query-efficient graph representation.

The first step is metadata ingestion and normalization. In this step, source-specific parsers are implemented to interpret the distinct metadata structures of individual catalogs and map them onto the unified ontology model. Then, a text embedding model is employed to compute high-dimensional semantic vector representations for the entities using their representative attributes, such as titles and descriptions of *Organization*, and names of *Keyword*. It is worth noting that *Topic* entities do not naturally exist in the metadata. Instead, they are automatically generated from dataset titles and descriptions using LLM, and corresponding vector embeddings are computed at the same time. Based on these generated topics, the knowledge graph undergoes a topic governance process to reduce duplication and improve topic quality.

Next is a post-processing step, the system further enriches missing spatial and temporal metadata in the knowledge graph. This step identifies datasets lacking explicit spatial or temporal entities and extracts relevant information from structured metadata fields and extended source APIs, and, when necessary, infers spatial or temporal extents from descriptive text using LLM. This enrichment process substantially improves the completeness of spatiotemporal coverage in the knowledge graph and enhances support for spatially and temporally constrained queries.

The final step focuses on index creation and performance optimization. Multiple types of indexes are created to support different query patterns, including vector indexes for semantic similarity search, spatial indexes for spatial queries, and temporal indexes for time-range filtering. All indexes can be implemented using native graph database mechanisms such as Neo4j indexing.

## 3.2 Multi-agent framework

### 3.2.1 Orchestration and memory module

The orchestration and memory module serves as the system-level core of the multi-agent geospatial data discovery framework. During runtime, the orchestration module models geospatial data discovery as a workflow composed of multiple interdependent subtasks. It not only triggers and coordinates the execution of specialized agents, but also monitors and evaluates their outputs, persisting critical state information to the memory module. Its core responsibilities include workflow orchestration, service coordination, state management, routing control, exception handling, and data persistence, ensuring robust system operation under complex queries and uncertain conditions.

The memory module provides structured short-term and long-term memory support, continuously storing key contextual information generated throughout the discovery process. This includes users' original queries and interaction histories, parsed and structured intent representations, and intermediate outputs produced by individual agents at different stages. Through unified management of problem-level and session-level memory, the system supports progressive evolution of user intent across multiple interaction rounds and supplies contextual grounding for subsequent retrieval, ranking, and result synthesis. In addition, the memory module preserves a complete execution trace of the discovery process, enabling transparency and reproducibility of system behavior and facilitating result interpretation and methodological evaluation.

### 3.2.2 Intent parsing agent

The intent parsing agent is the first core agent in the multi-agent intelligent geospatial data discovery pipeline. Its primary function is to transform user query expressed in natural language into structured geospatial intent representations (GeoIntent). Fig. 3 shows the workflow of the intent parsing agent.

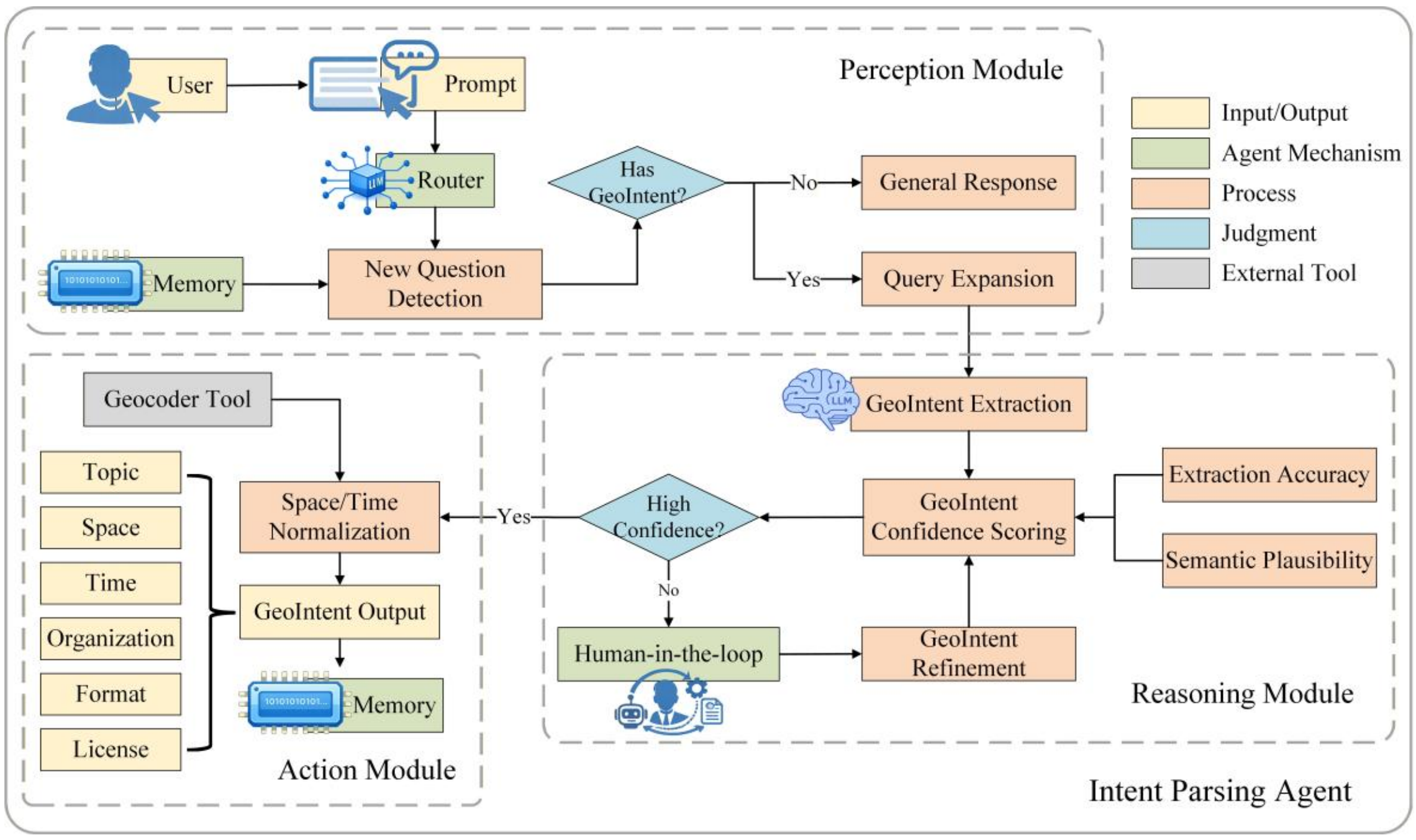

Fig. 3. Workflow of the intent parsing agent.

First, the agent employs the LLM to perform new question detection, utilizing session memory to determine whether the current input represents a new discovery question or a refinement of an existing one. This step enables the agent to distinguish between question evolution and question switching in multi-turn interaction scenarios, thereby avoiding redundant parsing or incorrect inheritance of prior intent states. A routing mechanism is then applied to assess whether the user query contains an explicit geospatial data discovery intent. Queries unrelated to data discovery are handled with general responses, whereas discovery-oriented queries are forwarded to following steps for full intent interpretation.

During the following intent parsing stage, the agent uses an LLM, when necessary, to perform query expansion and semantic rewriting to recover implicit conditions, resolve ambiguities, and improve semantic completeness. The agent then employs the LLM to extract key components of geospatial intent from the query, including thematic focus (*Topic*), spatial extent (*Space*), temporal extent (*Time*), as well as constraints of *Organization*, *Format*, and *License*.

To mitigate the risk of unreliable interpretations produced by LLM in complex or ambiguous query scenarios, a confidence assessment mechanism is introduced after intent extraction. This mechanism jointly considers extraction accuracy and semantic plausibility among intent components and uses an LLM to assign a confidence score to the parsed intent.

Formally, for a parsed intent $I$, the confidence score is defined as Eq. (1):

$$Conf(I) = \alpha \cdot A(I) + \beta \cdot P(I) \tag{1}$$

where $A(I) \in [0,1]$ denotes the extraction accuracy score, $P(I) \in [0,1]$ represents the semantic plausibility score, $\alpha$ and $\beta$ are weighting coefficients satisfying $\alpha + \beta = 1$.

When the confidence score exceeds a predefined threshold, the agent further normalizes spatial and temporal elements. Spatial descriptions are resolved using a geocoder tool to obtain standardized geographic representations, which are then converted into bounding boxes, while temporal expressions are normalized into unified time interval formats. The resulting GeoIntent representation can then be directly consumed by the graph retrieval stage and is simultaneously stored in the memory module.

When the confidence score falls below the threshold, a human-in-the-loop mechanism is triggered to guide users in clarifying query conditions or confirming key constraints. User feedback is recorded in the memory module and used to update the current session state, after which the intent parsing agent re-executes the parsing process based on the supplemented information.

### 3.2.3 Graph retrieval agent

The graph retrieval agent performs intent-driven retrieval and ranking within the structured geospatial metadata knowledge graph. Its primary objective is to efficiently identify, filter, and rank datasets that are most relevant to the user's intent from large-scale and heterogeneous datasets, while strictly satisfying explicit constraints. Fig. 4 shows the workflow of the graph retrieval agent.

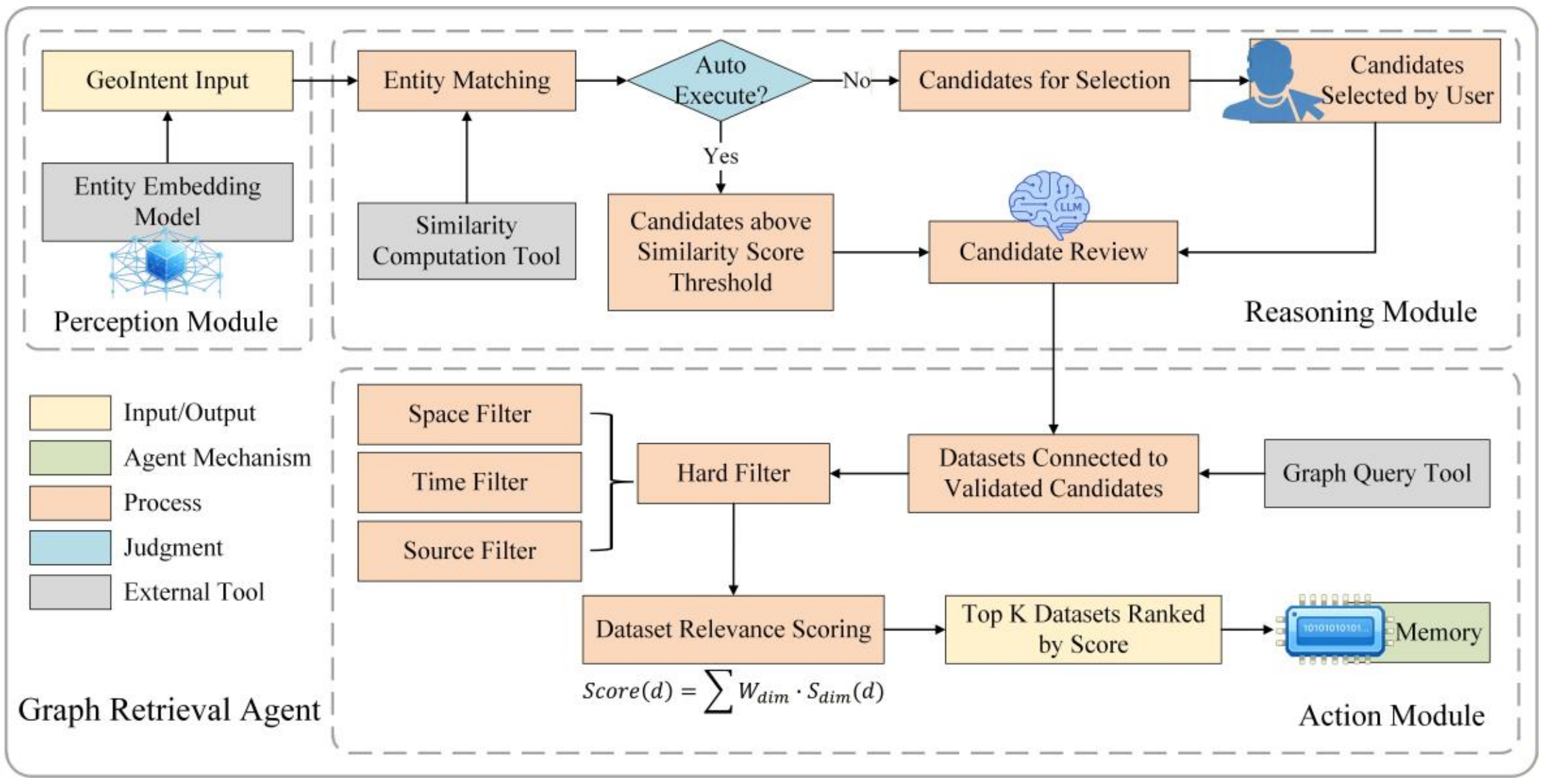

Fig. 4. Workflow of the graph retrieval agent.

The retrieval process begins with an entity matching stage. The GeoIntent is first transformed into high-dimensional vector representations by entity embedding model. The graph retrieval agent then maps the constraint elements contained in GeoIntent to their corresponding entity nodes in the knowledge graph through entity embedding similarity computation in vector space, identifying candidate entities that are semantically closest to the intended constraints. To balance automation efficiency and user control, the agent supports both automatic and manual execution modes. After candidate entities are determined through automatic threshold or user selection, the agent invokes an LLM-based semantic validation module to review the candidate set, ensuring that each entity is semantically consistent with the original GeoIntent and removing entities with spurious embedding similarity.

Once candidate entities are determined, the graph retrieval agent calls graph query tool to retrieve dataset nodes that are directly or indirectly connected to these entities, forming an initial candidate dataset set. The agent then applies hard filtering, enforcing non-negotiable constraints across dimensions of spatial extent, temporal coverage, and data source. Spatial filtering requires that the dataset's spatial coverage intersect with the user-specified spatial constraint, while temporal filtering requires overlap between the dataset's time span and the target time interval.

After hard filtering, the graph retrieval agent performs multi-dimensional relevance scoring on the remaining candidate datasets. The relevance score is computed using a weighted linear aggregation model that integrates multiple intent-related dimensions. Formally, for a candidate dataset $d$, the raw relevance score is defined as Eq. (2):

$$\begin{aligned} Score_{raw}(d) = W_{Topic} \cdot S_{Topic}(d) + W_{Space} \cdot S_{Space}(d) + W_{Time} \cdot S_{Time}(d) \\ + W_{Organization} \cdot S_{Organization}(d) + W_{Format} \cdot S_{Format}(d) \\ + W_{License} \cdot S_{License}(d) \end{aligned} \quad (2)$$

where $S_{dim}(d) \in [0,1]$ denotes the score of dataset $d$ along dimension $dim$, and $W_{dim}$ represents the corresponding weight. All weights are normalized such that $\sum W_{dim} = 1$.

For semantic dimensions of *Topic*, *Organization*, *Format*, and *License*, scores are derived from the maximum similarity between dataset-associated entities and candidate intent entities identified during entity matching. *Keyword* entities are incorporated into the *Topic* dimension as complementary semantic signals. Spatial and temporal scores are computed based on the overlap between the query constraints and dataset coverage. To balance coverage and precision, both dimensions adopt an F1 score-based metric defined as Eq. (3):

$$F1 = \frac{2 \cdot Precision \cdot Recall}{Precision + Recall} \quad (3)$$

where $Precision$ measures the proportion of dataset coverage overlapping the query region, and $Recall$ measures the proportion of the query region covered by the dataset.

In the final stage, raw scores are normalized across all candidate datasets to ensure that final relevance scores lie within the interval $[0,1]$. Based on the final normalized scores, the agent ranks candidate datasets and outputs the Top K results.

### 3.2.4 Answer synthesis agent

The answer synthesis agent is responsible for reranking the candidate datasets produced by the graph retrieval agent and transforming them into structured, interpretable, and user-oriented final responses. Fig. 5 shows the workflow of the graph retrieval agent.

The synthesis process begins with the Top K datasets ranked by the graph retrieval agent using relevance score. For each candidate dataset, the agent collects supporting evidence from both the graph retrieval agent and original knowledge graph, including matched entities corresponding to the user's GeoIntent, similarity scores and dimensional contributions computed during the retrieval stage.

After evidence preparation, the answer synthesis agent constructs a structured prompt that integrates the user's original query, the parsed GeoIntent representation, relevant conversation history, and the structured evidence associated with the Top K dataset. This design enables the LLM to generate explanatory text under explicit semantic context and clearly defined matching criteria. The generation process produces three primary outputs. First, based on the understanding capability of the LLM, the agent performs reranking over the Top K candidates to generate a final Top N list, where N≤K. Second, concise summaries are generated for the entire Top N datasets, highlighting core themes, spatial coverage, and temporal extent to facilitate rapid user comprehension. Third, explicit matching rationales are generated for each dataset, detailing how it satisfies the user's intent across thematic, spatial, temporal, and other constraint dimensions.

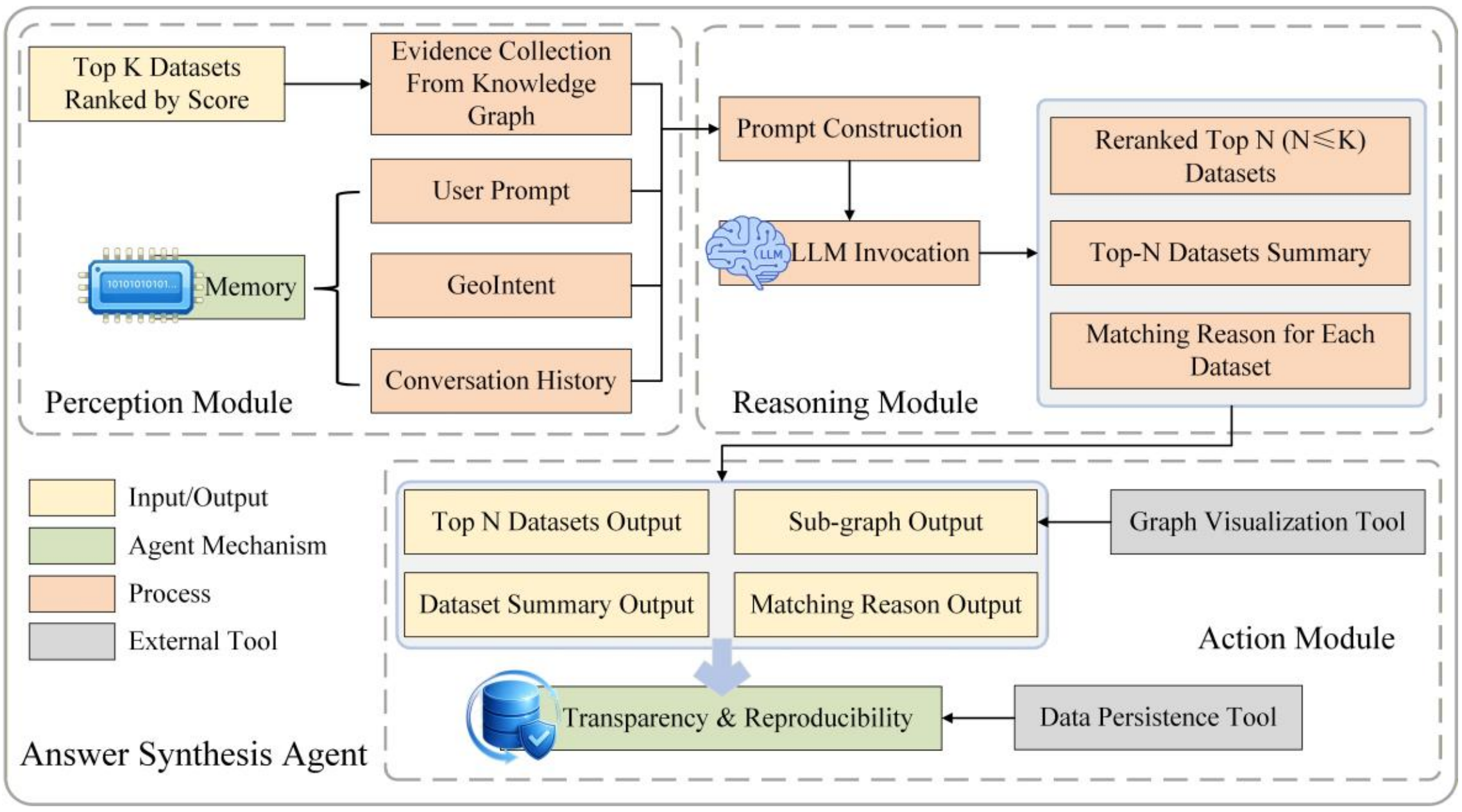

Fig. 5. Workflow of the answer synthesis agent.

To enhance transparency, the answer synthesis agent not only provides textual explanations but also generates associated subgraph structures that illustrate the connections between datasets and key entities within the knowledge graph. These structured subgraph outputs offer traceable semantic evidence, strengthening the transparency and reproducibility of the discovery results.

## 3.3 Evaluation

### 3.3.1 Performance evaluation indicators

To quantitatively assess the effectiveness of the intelligent geospatial data discovery framework, this study adopts three indicators for the performance evaluation. First, EIMR (Exact Intent Match Rate) is used to evaluate the performance of the intent parsing agent. For each query $q$, let $S_q$ denotes the manually annotated ground-truth intent constraints (*Topic*, *Space*, *Time*, *Organization*, *Format*, and *License*) and $\hat{S}_q$ denote the intent constraints identified by the system. EIMR measures the proportion of correctly identified constraints within query $q$. EIMR is defined as Eq. (4):

$$EIMR(q) = \frac{|S_q \cap \hat{S}_q|}{|S_q|} \tag{4}$$

where $|S_q|$ represents total number of ground-truth constraints for query $q$, and $|S_q \cap \hat{S}_q|$ denotes the number of correctly identified constraints.

Second, NDCG@10 (Normalized Discounted Cumulative Gain at rank 10) evaluates the quality of ranking within the top 10 returned results. It accounts for both the relevance of retrieved items and their ranking positions, assigning higher weight to results appearing at higher ranks.

Before NDCG@10, the DCG@10 (Discounted Cumulative Gain at rank 10) is defined as Eq. (5):

$$DCG@10 = \sum_{i=1}^{10} \frac{rel_i}{\log_2(i+1)} \quad (5)$$

where $rel_i$ denotes the graded relevance score of the item at rank $i$. To normalize this value and allow comparison across queries with different relevance distributions, DCG@10 is divided by the IDCG@10 (Ideal DCG at rank 10), which forms NDCG@10 represents the maximum possible DCG under perfect ranking. NDCG@10 is defined as Eq. (6):

$$NDCG@10 = \frac{DCG@10}{IDCG@10} \quad (6)$$

NDCG@10 ranges from 0 to 1, where higher values indicate that highly relevant items are ranked closer to the top of the result list.

Third, Recall@20 measures the retrieval coverage within the top 20 returned results. It reflects the system's ability to retrieve relevant datasets among all relevant items identified for a given query in the pool. Recall@20 is defined as Eq. (7):

$$Recall@20 = \frac{|\{relevant\ items\ in\ Top\ 20\}|}{|\{all\ relevant\ items\ for\ the\ query\ in\ the\ pool\}|} \quad (7)$$

Recall@20 evaluates how many relevant datasets are successfully retrieved within the top 20 positions, regardless of their exact ranking order.

### 3.3.2 Evaluation design

The evaluation is designed to ensure fairness, reproducibility, and methodological rigor, based on a pre-constructed set of 100 test queries including diverse geographic domains. To systematically examine retrieval performance under varying levels of query complexity, the 100 queries are divided equally into four hierarchical levels according to constraint complexity:

(1) Level 1 (topical queries): Queries containing only a thematic constraint.
(2) Level 2 (single spatiotemporal constraint queries): Queries including a thematic constraint together with either a spatial or a temporal constraint.
(3) Level 3 (spatiotemporal queries): Queries incorporating a thematic constraint along with both spatial and temporal constraints.
(4) Level 4 (complex queries): Queries containing thematic, spatial, and temporal constraints in addition to one or more supplementary constraints (e.g., organization, format, and/or license).

For NDCG@10 and Recall@20, a unified pooled datasets should be collected, merged and deduplicated from different retrieval components requiring comparison. All subsequent relevance judgments and metric computations are performed based on it, ensuring that comparisons across systems are grounded on a consistent evaluation basis.

Relevance assessment adopts a graded scale from 0 to 3 to reflect varying degrees of alignment between retrieved datasets and query intent. The scoring rubric is defined as follows:

(1) 3 (highly relevant): The dataset fully satisfies the core topic and all explicit constraints and can be directly used for downstream tasks.
(2) 2 (substantially relevant): The dataset satisfies the core topic and most key constraints, with some deviations, and remains practically useful.

(3) 1 (marginally relevant): The dataset is related to the topic but does not satisfy important constraints or only partially aligns with the intended scope.

(4) 0 (not relevant): The dataset does not meaningfully match the topic or constraints of the query, exhibiting only superficial keyword overlap or weak semantic alignment.

Two doctoral researchers with backgrounds in geographic information science collaboratively evaluate all pooled datasets. Prior to scoring, the evaluators receive standardized training based on a unified scoring guideline. Evaluators are instructed to base their judgments solely on the query and dataset metadata content, without considering ranking position or system source.

For the computation of Recall@20, graded relevance scores are binarized using a predefined threshold: datasets with scores above 0 are considered relevant, while those with score 0 are considered non-relevant. In contrast, NDCG@10 directly incorporates the original 0-3 graded relevance scores without binarization.

## 4. Result

### 4.1 Prototype system

To validate the feasibility and effectiveness of the proposed framework, we developed a prototype system to implement the complete intelligent geospatial data discovery workflow (Fig. 6).

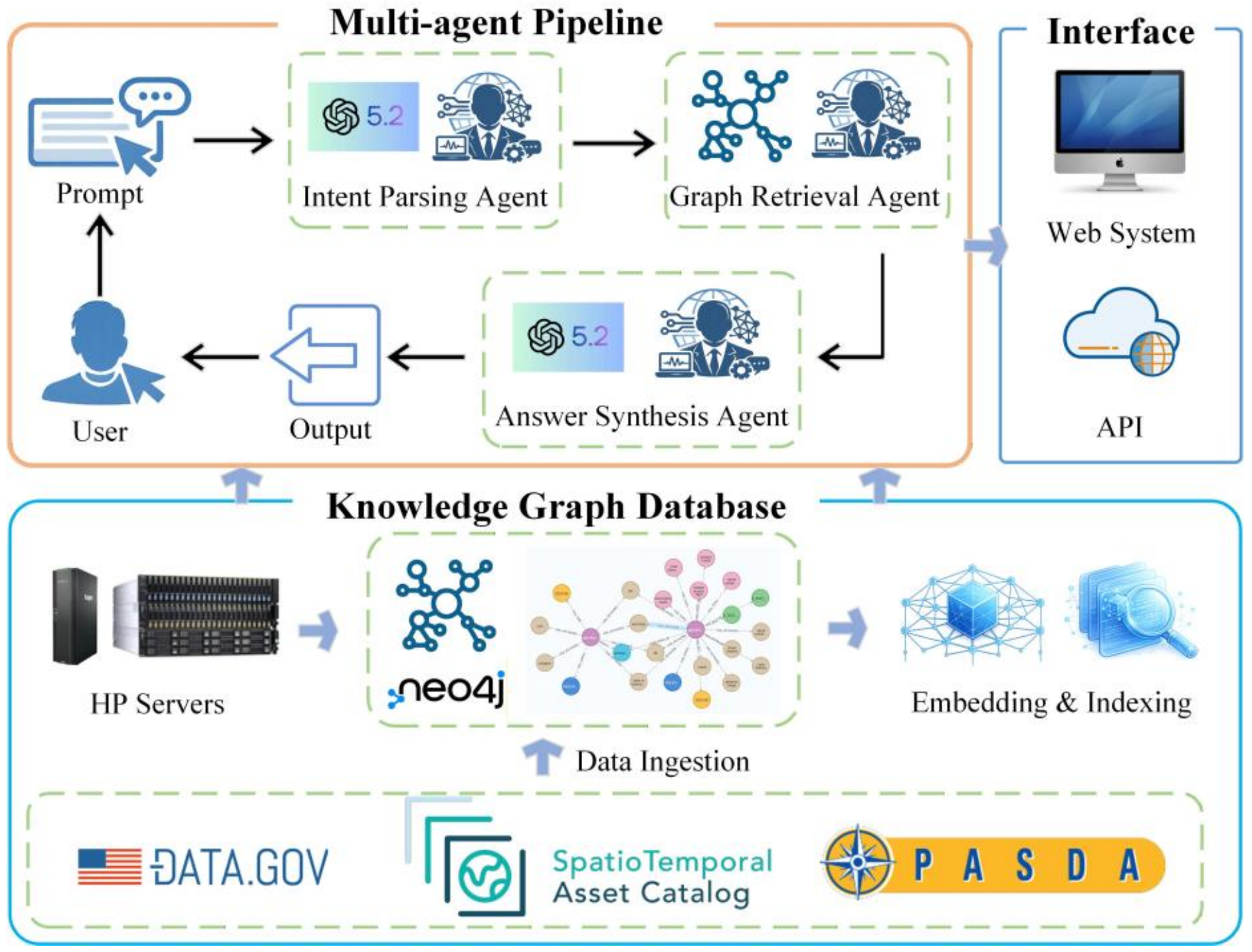

Fig. 6. Prototype system architecture of intelligent geospatial data discovery (IGDD).

We deployed the knowledge graph on Neo4j within a high-performance server environment. Neo4j is a native graph database platform that supports the property graph model, vector indexing, and spatial operations (Vukotic et al., 2014). The graph schema follows the proposed geospatial metadata ontology and integrates six data catalogs with 264,177 datasets, including four STAC endpoints, Pennsylvania Spatial Data Access (PASDA) following FGDC CSDGM standard, and

Data.gov conforming to CKAN, as summarized in Table 1. For the implementation details, we used the text-embedding-3-large model from OpenAI for entity embeddings and created vector indexes for entities. Spatial and temporal indexes were also established for corresponding entities to improve filtering and retrieval performance. After construction, the knowledge graph contains 2,796,946 entities and 12,229,077 relationships.

Table 1. Data catalogs that are integrated into the knowledge graph.

| **Data Catalog** | **Dataset Count** | **Metadata Standard** | **Access Link** | **API Link** |
|---|---|---|---|---|
| Google Earth Engine Proxy for openEO | 940 | STAC | https://stacindex.org/catalogs/google-earth-engine-openeo | https://earthengine.openeo.org/v1.0/ |
| Microsoft Planetary Computer STAC API | 126 | STAC | https://stacindex.org/catalogs/microsoft-pc | https://planetarycomputer.microsoft.com/api/stac/v1/ |
| Destination Earth Data Lake Harmonized Data Access (DEDL HDA) STAC API | 199 | STAC | https://stacindex.org/catalogs/dedl-api | https://hda.data.destination-earth.eu/stac/v2 |
| Paituli STAC with Finnish data | 177 | STAC | https://stacindex.org/catalogs/paituli-stac-finland | https://paituli.csc.fi/geoserver/ogc/stac/v1 |
| Pennsylvania Spatial Data Access (PASDA) | 2,083 | FGDC CSDGM | https://www.pasda.psu.edu/ | https://www.pasda.psu.edu/uci/DataSummary.aspx?dataset={id} |
| Data.gov (only geospatial type) | 260,652 | CKAN | https://data.gov/ | https://catalog.data.gov/api/3/action/package_search?fq=metadata_type:geospatial |

For the multi-agent pipeline, the system integrated the GPT-5.2 API from OpenAI as the underlying LLM service. At the frontend level, the prototype provides a conversational interface that allows users to submit natural language queries and receive structured discovery results. As shown in Fig. 7, the left panel presents user configuration options and conversation history. The central panel functions as the interactive dialogue interface. The right panel presents pipeline execution status, parsed intent representations, and system logs, offering transparency throughout the discovery workflow.

For the subsequent experiments, the system operated under the automatic execution mode by default. In this mode, default configuration parameters were adopted as follows: $W_{Topic} = 0.3$, $W_{Organization} = 0.1$, $W_{Space} = 0.2$, $W_{Time} = 0.2$, $W_{Format} = 0.1$, $W_{License} = 0.1$. The confidence threshold of intent parsing was set to 0.5, and the similarity score threshold for entity embedding matching was set to 0.7. In the graph retrieval stage, the default Top K candidate set was fixed at 20, and the final Top N results presented to users were set to 10.

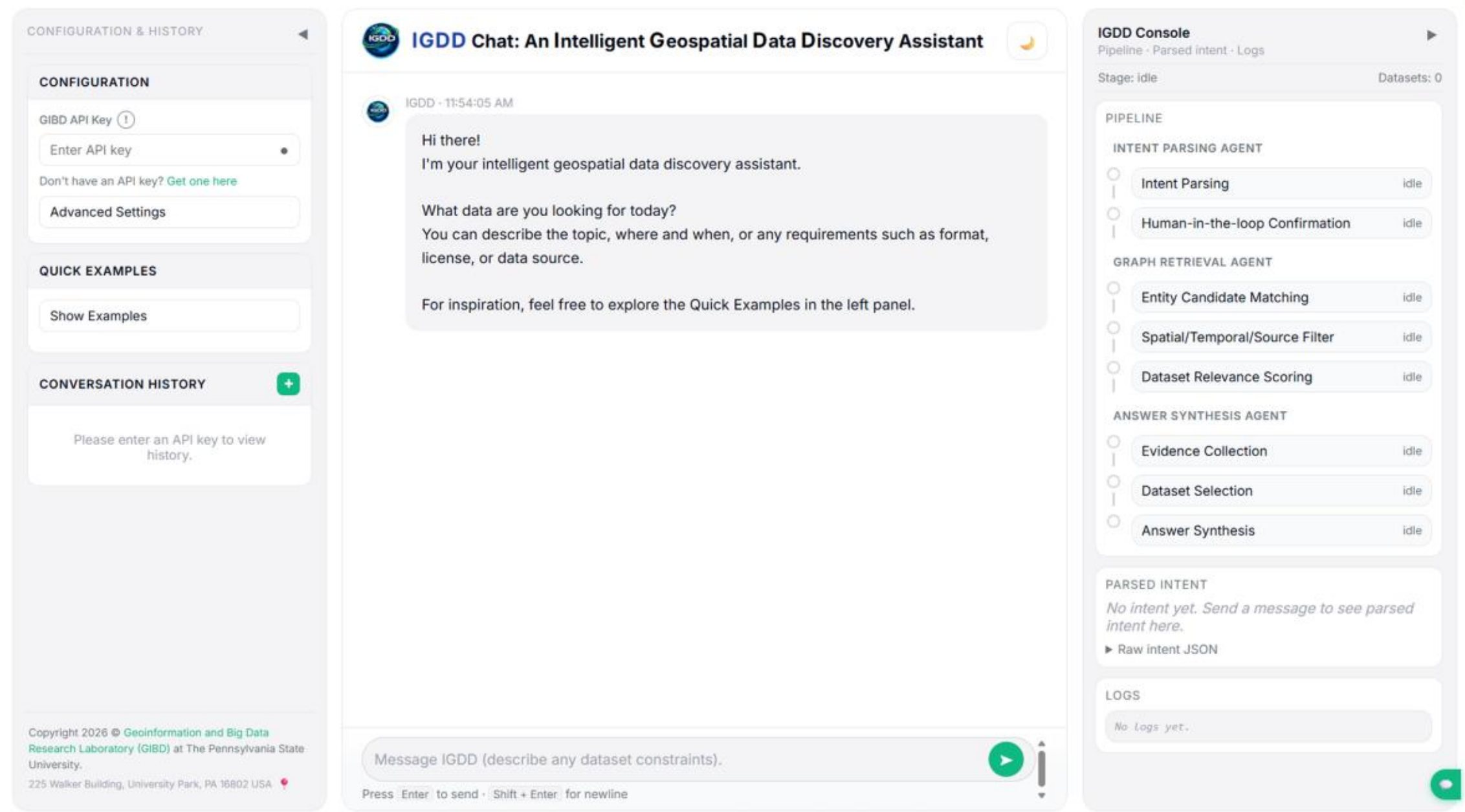

Fig. 7. Graphic user interface of the intelligent geospatial data discovery system.

## 4.2 Representative use cases

### 4.2.1 Climate data discovery

This case shows how the system helps researchers find long-term daily temperature data for trend or extreme climate studies. For example, a researcher may ask: "I'm looking for daily temperature datasets for CONUS from 1990 to 2020." This query is brief but clearly states the topic, space, and time constraints.

First, the intent parsing agent performed new question detection to confirm that it represents a new discovery question and conducted semantic parsing and normalization. The phrase "daily temperature" was identified as the thematic constraint, while "CONUS" was recognized as the spatial constraint and normalized into a standardized continental United States bounding box [-124.7844, 24.3963, -66.9514, 49.3844]. The temporal phrase "from 1990 to 2020" was converted into a standardized time interval [19900101 00:00:00, 20201231 23:59:59]. The confidence scores for *Topic*, *Space*, and *Time* extraction were 90%, 78%, and 90%, respectively. Since all scores exceeded the predefined threshold 0.7, the system did not trigger the human-in-the-loop stage (Figure 8a).

During the graph retrieval stage, the embedding of thematic constraint "daily temperature" was used to search all *Topic* entities in the knowledge graph. Entities with similarity scores above 70% were retained, including "daily air temperature" (89.25%), "daily temperature normals" (89.06%), "daily temperature composites" (85.18%), etc. The system then performed graph query to retrieve datasets associated with these *Topic* entities. Hard constraint filtering was subsequently applied to retain only datasets whose spatiotemporal coverage intersects with the constraints. After candidate datasets were identified, multidimensional relevance scoring was performed based on embedding similarity or F1 measure, followed by normalization. The system then produced a ranked Top 20 dataset list, as shown in the Appendix Table. B1.

It can be observed that the retrieved candidates originated primarily from Google Earth Engine Proxy for openEO and Data.gov. Due to the relatively high semantic similarity between "daily temperature" and related concepts such as "monthly temperature", certain noisy datasets were included in the Top 20 results, for example the "PRISM Monthly Spatial Climate Dataset ANm". In addition, unrelated datasets such as "SST, Aqua MODIS, NPP, 0.0125°, West US, Day time (11 microns), 2002-present (1 Day Composite)", which represent sea surface temperature products rather than continental air temperature datasets, also appeared in the initial ranking. These cases illustrate the limitations of purely similarity- and weight-based ranking mechanisms.

The workflow then proceeded to the answer synthesis agent. Structured evidence for each candidate dataset was extracted from the knowledge graph, and the memory module provided the original user prompt, conversation history, and structured GeoIntent. These elements were incorporated into a structured prompt and passed to the LLM which reranked the Top 20 candidates and selected the final Top 10 datasets while generating explanatory output. Notably, the LLM successfully eliminated all noisy datasets, as presented by the reranking number in the "Synthesis Ranking by LLM" column in Appendix Table. B1. Utilizing the outputs from answer synthesis agent, the web system presented the results for users (Fig. 8).

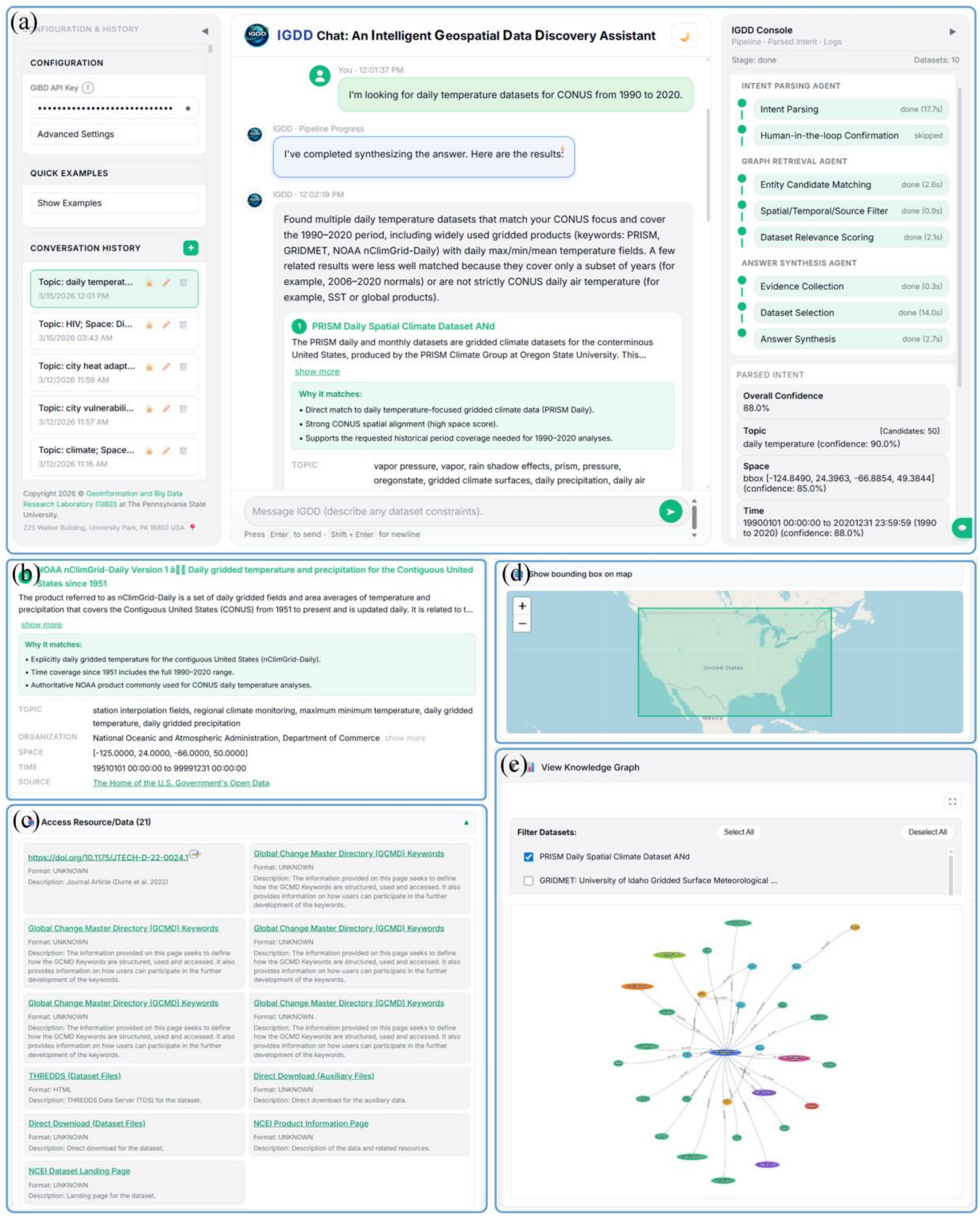

Fig. 8. Results shown in the web system for climate data discovery: (a) Main interface view; (b) The third ranked dataset after scrolling down; (c) The resource entities associated with the third dataset; (d) Visualization of the bounding box of the third dataset; (e) A subgraph visualization corresponding to the selected dataset. (video demonstration)

### 4.2.2 U.S. census data discovery

The second use case concerns U.S. Census data that support researchers or planners in obtaining road datasets for a specific region and year. Suppose a user submitted the following query: "Please find primary roads datasets provided by the U.S. Census Bureau for California for the year 2022, available in ZIP format." This query explicitly specifies five constraints: topic, space, time, organization, and format.

Following the standard workflow, the intent parsing agent identified "primary roads", "U.S. Census Bureau", "California", "2022", and "ZIP" as the constraints. During the graph retrieval stage, entity matching and multidimensional relevance scoring was performed. Subsequently, the answer synthesis agent reranked the Top 20 candidates and generated the final Top 10 results, as shown in Appendix Table. B2**Error! Reference source not found.**.

The results demonstrated that the answer synthesis agent successfully identified the most relevant dataset and ranked it in the first position. The second-ranked dataset corresponds to nationwide data that can be clipped to the California region. Rankings three through six consist of county-level datasets within California, which also satisfy the query requirements. Rankings seven through ten correspond to datasets from neighboring states such as Nevada, Oregon, Idaho, and Arizona, whose bounding boxes intersect California. Fig. 9 presents part of the results shown in the web system.

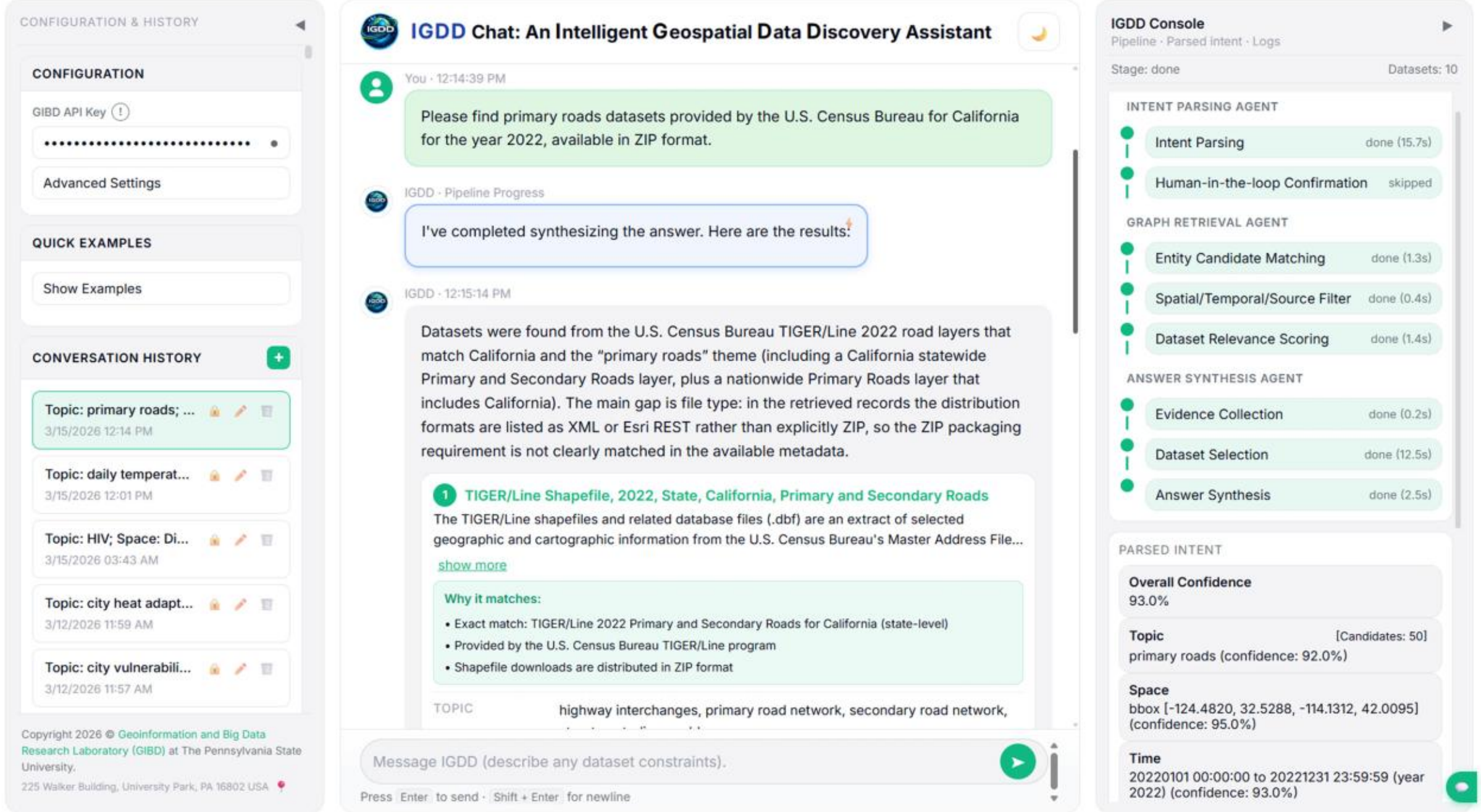

Fig. 9. Part of the results shown in the web system for U.S. census data discovery. (video demonstration)

### 4.2.3 Public health data discovery

To evaluate the effectiveness of the proposed framework in public health data discovery tasks, we selected two representative health-related queries covering HIV and opioid use. To provide a preliminary comparison with Data.gov, the source of the system was restricted to Data.gov.

For the first query "I am looking for HIV data in District of Columbia.", the identified datasets "HIV and AIDS Clinic" and "AIDS Drug Assistance Program" both demonstrate direct thematic alignment with HIV/AIDS-related content, achieving high topic similarity scores (0.94). The dataset “DC Health Planning Neighborhoods to Census Tracts” can support spatial aggregation and mapping of HIV metrics within DC health planning geographies although it does not explicitly contain HIV indicators. In contrast, when using Data.gov with the keyword “HIV” (Fig. 10a), only “HIV and AIDS Clinic” and “AIDS Drug Assistance Program” were retrieved after relevance ranking. This comparison indicates that while both systems identified directly matching datasets, the proposed framework further identified an auxiliary dataset beneficial for downstream spatial analysis.

For the second query “Please find opioid use datasets published by the City of Tempe, licensed under the Creative Commons Attribution license, and provided in GeoJSON format.”, the Top 3 returned datasets all exhibit strong thematic alignment with opioid use while satisfying all other constraints, which can also be proved by matching reasons. In comparison, when querying Data.gov (Fig. 10b), only “Opioid EMS Calls” and “1.21 Youth Drug Use and Misuse (summary)” were returned. Upon inspection, the latter dataset is not directly related to opioid use. This comparison highlights that the proposed system achieves stronger relevance and higher recall of truly relevant datasets than traditional data discovery. Appendix Table. B3 presents the top three discovery results for each query.

Fig. 10. Public health data discovery results in Data.gov for comparison: (a) HIV; (b) opioid use.

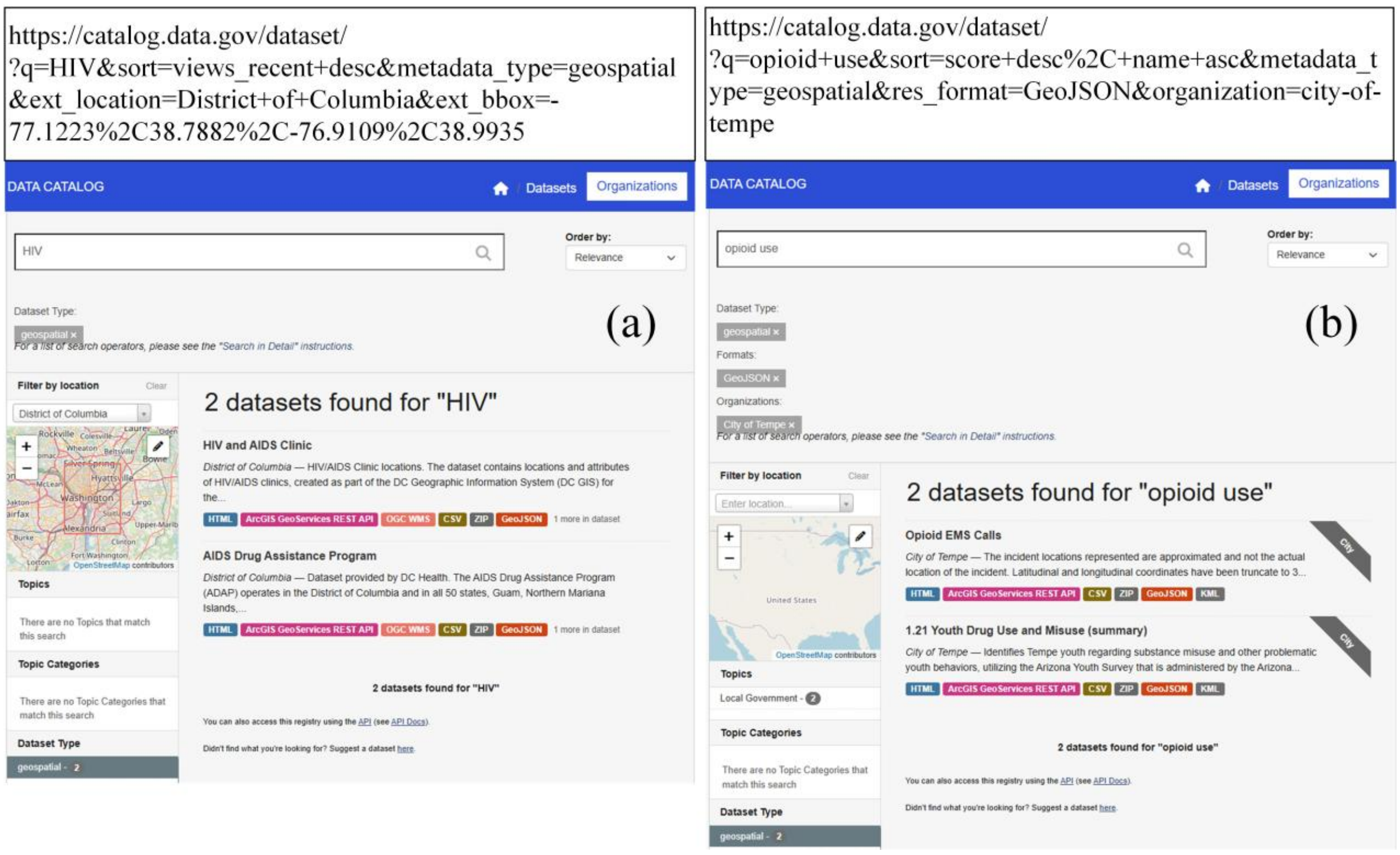

### 4.2.4 Satellite imagery data discovery

To evaluate the system’s capability in satellite imagery data discovery, this study constructed comparative scenarios centered on two representative raster products: DEM (Digital Elevation Model) and NDVI (Normalized Difference Vegetation Index) at two spatial scales: global coverage and national coverage for Finland. Fig. 11 presents the ranked discovery results and their corresponding data sources.

At the global scale, the system performed collaborative discovery and integration across multiple heterogeneous data sources. The returned DEM datasets originated from diverse platforms, including Data.gov, the DEDL HDA STAC API, and the Google Earth Engine Proxy for openEO. Global DEM products include ETOPO, SRTM, Copernicus DEM, and MERIT DEM, among others. Similarly, global NDVI results encompass multi-source time-series products such as NOAA CDR AVHRR NDVI, GIMMS NDVI, and the MODIS Terra and Aqua series. These results demonstrated that the system integrates heterogeneous data sources within a unified semantic framework, thereby enhancing the completeness and coverage of global-scale data discovery.

At the national scale for Finland, the majority of DEM and NDVI datasets covering Finland were retrieved from the Paituli STAC platform and its associated GeoCubes services. The top-ranked results include high-resolution elevation models at 2 m, 10 m, and 25 m resolutions, as well as various NDVI and terrain-derived products generated within Finland's national data production infrastructure. This result structure indicates that, after integrating the Paituli STAC catalog, the system prioritizes authoritative and regionally well-matched datasets for national-scale queries. At the same time, when region-specific resources are limited, the system retains relevant global datasets as complementary candidates, forming a multi-layered result set.

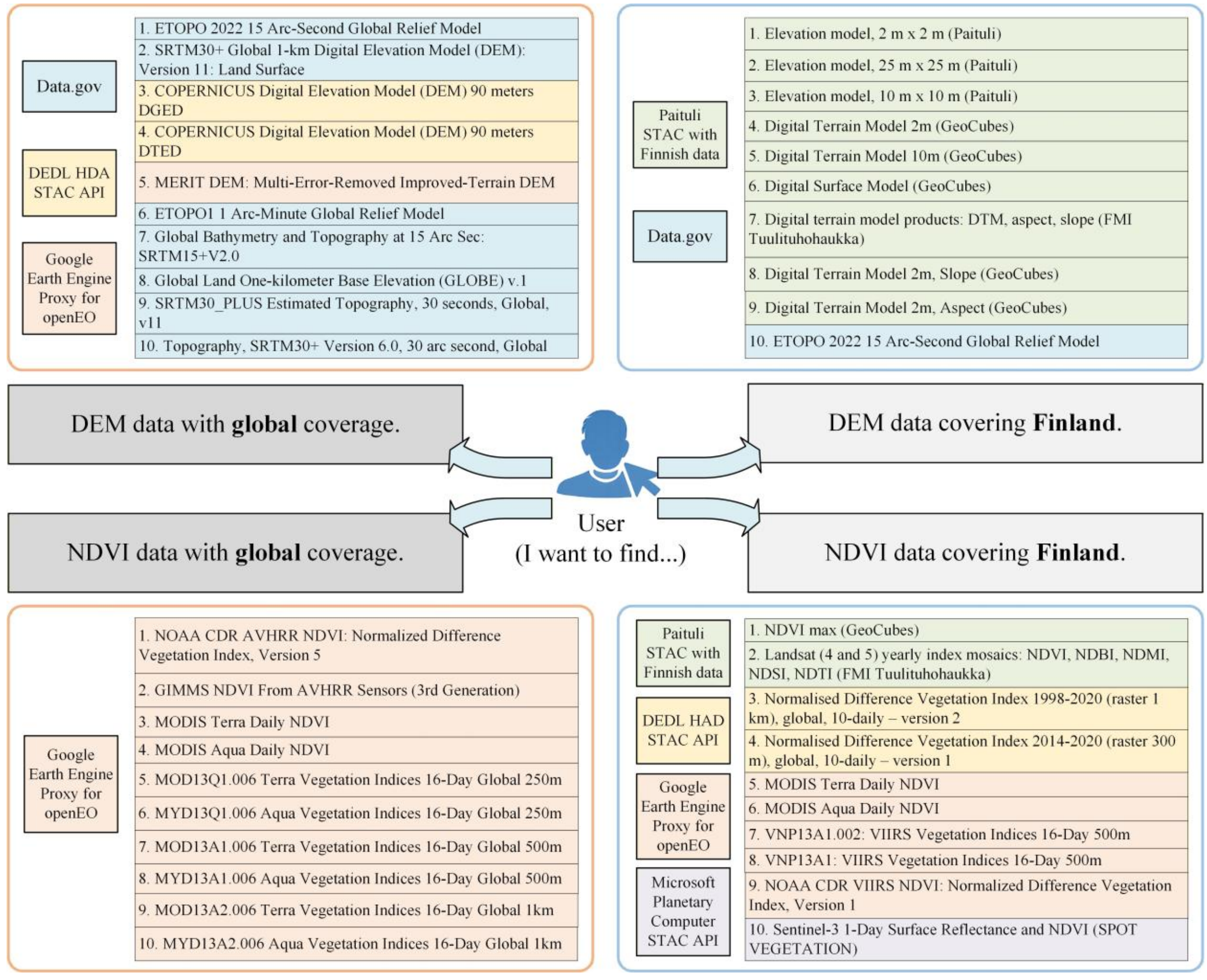

Fig. 11. Satellite imagery data discovery results (video demonstration)

By combining global and national-scale satellite imagery data discovery scenarios, this case study demonstrates the proposed framework's capability in multi-source collaboration, semantic harmonization, and spatially precise matching. Together, these results validate the system's adaptability and effectiveness across different spatial scales.

### 4.2.5 Local data catalog discovery

To evaluate the adaptability and practical applicability of the proposed framework in local data catalogs, this study employs PASDA, the official geospatial data portal of the Commonwealth of Pennsylvania, as a demonstration case for local data catalog integration and discovery. PASDA aggregates a wide range of statewide thematic datasets, including Imagery, Lidar and Elevation, Hydrology, Transportation, Biota, Energy, etc.

During the discovery process, when users submit queries related to Pennsylvania, the system prioritizes and retrieves highly relevant datasets from PASDA. Fig. 12 illustrates four representative datasets successfully discovered from PASDA. These datasets encompass multiple geometric types, including point, line, polygon, and raster formats, demonstrating the system's ability to uniformly support diverse spatial data types within a single semantic discovery framework.

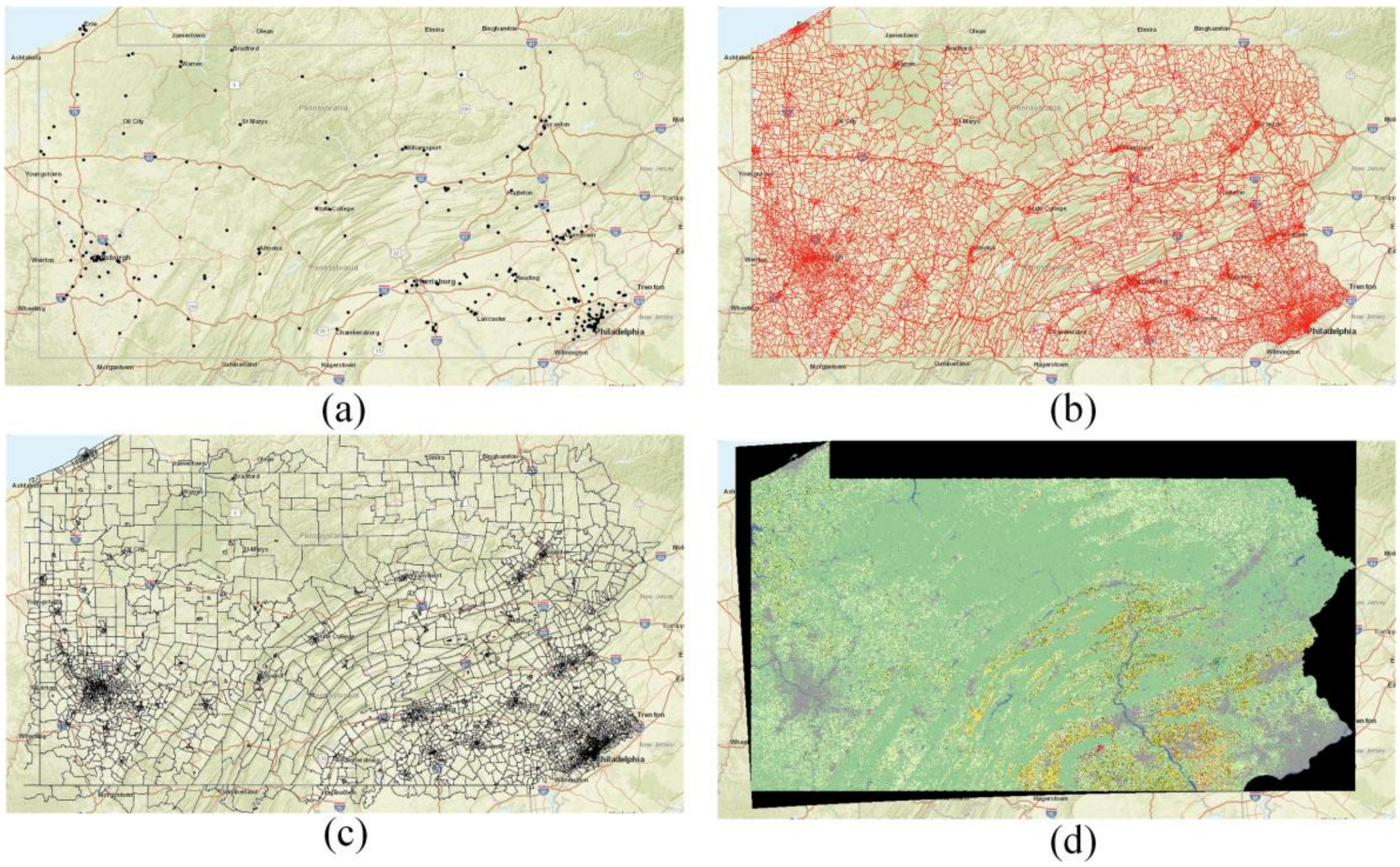

Fig. 12. Representative datasets discovered from PASDA: (a) hospital point; (b) road network; (c) census block group boundary; (d) cropland.

Furthermore, the system not only performs collaborative discovery across multiple data sources but also implements a "local-first" strategy at the regional scale. When relevant datasets are available within PASDA, the system prioritizes these authoritative regional resources. If coverage within the local catalog is insufficient for a given topic, the system supplements result with relevant datasets from other sources, forming a multi-layered candidate set.

Through the PASDA local catalog demonstration, the scalability and practical utility of the proposed framework at the regional scale become evident. It supports complex regional analysis needs across multiple domains, including infrastructure, demographic units, and land use. This demonstration confirms that the framework is also capable of delivering precise and efficient data integration and discovery within localized data ecosystems.

### 4.2.6 Human-in-the-loop and multi-turn conversation demonstration

To demonstrate the system's capability in human-in-the-loop interaction and multi-turn conversation, this study designs a scenario centered on soil moisture, as illustrated in Fig. 13. At the initial stage of the conversation, the user submits a vague request: "I want to find soil moisture data in the South." During intent parsing, the system correctly identifies the topic as soil moisture but detects only an imprecise spatial reference, "South," with relatively low confidence (44%). Since "South" may correspond to multiple geographic contexts (e.g., the southern United States, the southern world), the system determines that the spatial constraint is ambiguous. It therefore activates the human-in-the-loop mechanism and proactively requests a more precise location specification, thereby avoiding the return of large volumes of uncertain or potentially irrelevant datasets.

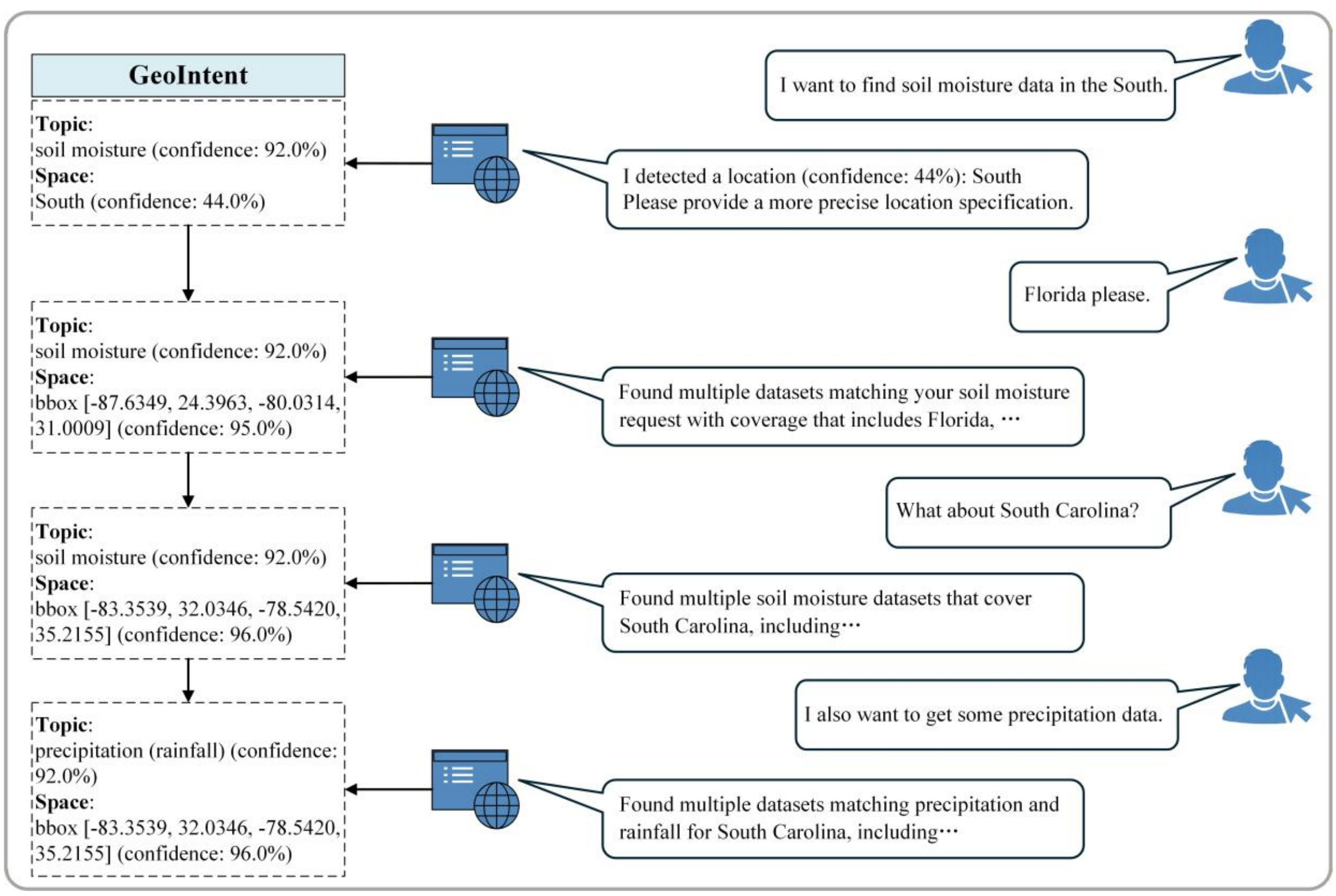

Fig. 13. Multi-turn conversation flow integrating human-in-the-loop.

After the user clarifies with "Florida please," the system updates the GeoIntent representation and resolves the spatial constraint to the bounding box coordinates corresponding to the state of Florida. Based on this refined intent structure, a new discovery process is executed, and multiple soil moisture datasets covering Florida are returned. At this stage, the user's intent has evolved from a vague regional description to a clearly defined administrative unit.

Subsequently, the user asks, “What about South Carolina?” The system performs new-question detection and determines that this is not an entirely new query but rather a modification of the existing one. Consequently, it does not reconstruct the full semantic structure from scratch. Instead, utilizing the current conversational context stored in memory, the system updates only the spatial constraint to South Carolina and re-executes the discovery process.

In the final stage of the conversation, the user further extends the request by stating, “I also want to get some precipitation data.” The system recognizes that the topic has shifted from soil moisture to precipitation and semantically expands precipitation to precipitation (rainfall) while maintaining the spatial constraint. Overall, this demonstration highlights the role of the human-in-the-loop mechanism in disambiguation and the role of multi-turn conversation in contextual continuity.

## 4.3 Performance evaluation results

This study conducted relevance pooling over 100 queries, resulting in a total of 3,646 pooled datasets. Relevance assessments were performed based on each query and its corresponding pooled set. Detailed information regarding the pool set and annotation procedure is provided in the data and codes availability statement section. The evaluation results are summarized as follows.

First, regarding the intent parsing agent, the overall EIMR is 99.27%. Across the four query complexity levels, the EIMR values were 100%, 100%, 98.68%, and 98.4%, respectively. These results demonstrate extremely high accuracy, indicating that the system is capable of reliably identifying and structuring user intent even for the most complex queries.

In terms of retrieval and ranking quality, we compared NDCG@10 under different components (Table 2), including the ablation experiment between agents and comparison with Data.gov. The full system (IGDD-G&A) achieves the best performance with an NDCG@10 of 85.22%, substantially outperforming Data.gov (43.65%) by 41.57%. Even when restricted to Data.gov as the only source, IGDD(D)-G&A still reaches 74.29%, improved by 30.64%. The ablation results further show that removing the answer synthesis agent reduces performance from 85.22% to 68.40% in the full setting and from 74.29% to 59.52% in the Data.gov-only setting. This demonstrates the important role of the answer synthesis agent in improving the ranking quality of relevant datasets, which improves NDCG@10 by 16.82% for IGDD-G and 14.77% for IGDD(D)-G. Overall, the coordinated multi-agent architecture significantly enhances geospatial data discovery and ranking effectiveness compared with traditional portal-based search.

Table 2. Different system components used in the performance evaluation.

| Component | Description | NDCG@10 |
|---|---|---|
| IGDD-G&A | Intelligent geospatial data discovery system with intent parsing agent, graph retrieval agent, and answer synthesis agent. | 85.22% |
| IGDD(D)-G&A | Intelligent geospatial data discovery system with intent parsing agent, graph retrieval agent, and answer synthesis agent, while using Data.gov as the only source. | 74.29% |
| IGDD-G | Intelligent geospatial data discovery system with intent parsing agent and graph retrieval agent. | 68.40% |

| IGDD(D)-G | Intelligent geospatial data discovery system with intent parsing agent and graph retrieval agent, while using Data.gov as the only source. | 59.52% |
|---|---|---|
| Data.gov | Data.gov data portal. | 43.65% |

Fig. 14 (a) illustrates performance across different query complexity levels. While NDCG@10 of IGDD-G&A decreases slightly as query complexity increases, the decline for Data.gov is much more pronounced, indicating its limited capability in handling complex, multi-constraint requests. Across all levels, IGDD-G&A consistently outperforms both IGDD(D)-G&A and Data.gov. The ablation experiments in different query complexity levels also demonstrate substantial additional gains with the inclusion of the answer synthesis agent.

In addition, Recall@20 was compared between IGDD(D)-G&A and Data.gov. IGDD-G&A is not included in this comparison because it operates over a broader set of sources, resulting in a different evaluation baseline. Since the answer synthesis agent does not alter the set of retrieved datasets, the Recall@20 values of IGDD(D)-G&A and IGDD(D)-G are the same. Because the total number of relevant datasets for a query may exceed 20, Recall@20 should be interpreted comparatively rather than as an absolute measure. Overall, IGDD(D)-G&A achieves a Recall@20 of 66.54%, whereas Data.gov achieves 34.61%, representing an improvement of 31.93%. Fig. 14 (b) shows that the relative recall advantage of IGDD(D)-G&A over Data.gov increases with query complexity, reaching 60.64% at Level 4.

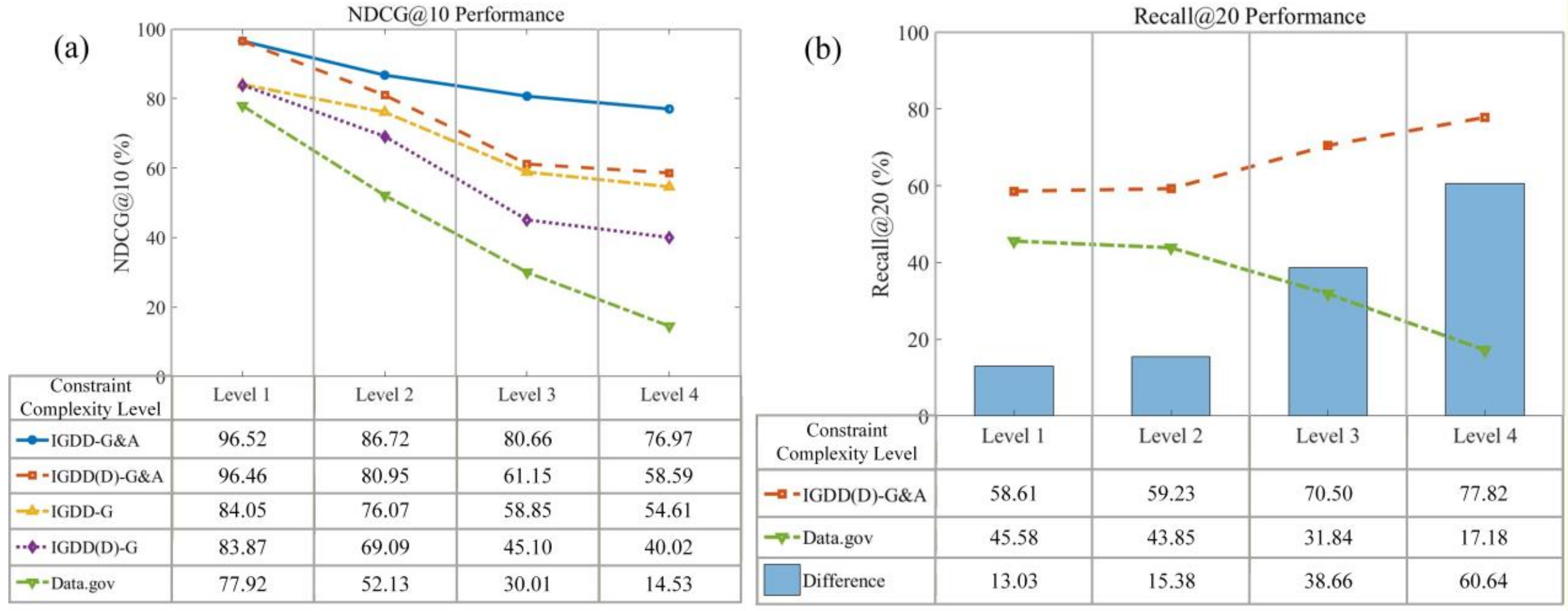

| Constraint Complexity Level | Level 1 | Level 2 | Level 3 | Level 4 |
|---|---|---|---|---|
| IGDD-G&A | 96.52 | 86.72 | 80.66 | 76.97 |
| IGDD(D)-G&A | 96.46 | 80.95 | 61.15 | 58.59 |
| IGDD-G | 84.05 | 76.07 | 58.85 | 54.61 |
| IGDD(D)-G | 83.87 | 69.09 | 45.10 | 40.02 |
| Data.gov | 77.92 | 52.13 | 30.01 | 14.53 |

| Constraint Complexity Level | Level 1 | Level 2 | Level 3 | Level 4 |
|---|---|---|---|---|
| IGDD(D)-G&A | 58.61 | 59.23 | 70.50 | 77.82 |
| Data.gov | 45.58 | 43.85 | 31.84 | 17.18 |
| Difference | 13.03 | 15.38 | 38.66 | 60.64 |

Fig. 14. Performance evaluation results under diverse constraint complexity levels: (a) NDCG@10; (b) Recall@20.

## 5. Discussion

A key strength of this framework lies in its scalability across heterogeneous data environments. The geospatial metadata ontology and knowledge graph developed in this study improve interoperability across catalogs and create a pathway for extending the framework to additional platforms and metadata ecosystems. From a system perspective, computational performance also remains practical. With caching strategies, vector index optimization, and graph database query optimization, graph-based retrieval over a knowledge graph containing more than 260,000 datasets typically requires only 1 to 2 seconds. When full end-to-end processing is considered, including GPT-5.2 latency, a complete discovery cycle generally takes about 30 to 40 seconds per query and consumes around 11,000 tokens. These results suggest that the framework is computationally feasible for real-world intelligent discovery tasks, although further gains in efficiency will remain important for larger-scale deployment.

Autonomous GIS emphasizes the ability of intelligent systems to complete the full workflow of data discovery, processing, analysis, and visualization with minimal human intervention. Within this vision, data discovery is the starting point of the autonomous analytical pipeline and a core prerequisite for downstream tasks. In this sense, the intelligent geospatial data discovery framework proposed in this study can be understood as a concrete step toward autonomous GIS, particularly the level 3 data-aware GIS, where the system is able to discover and access relevant data from online or local sources.

Despite these advances, several limitations also point to future research needs. First, this study focuses mainly on the data discovery stage. In practical scientific research and decision-making workflows, however, data preprocessing and the generation of analysis-ready data are equally important. Future work could extend the framework beyond discovery to encompass a complete pipeline from data discovery to data preprocessing. Second, geospatial datasets exhibit substantial variation in spatial and temporal resolution. The current framework does not fully address issues related to resolution matching and scale compatibility. Future research could introduce hierarchical resolution modeling mechanisms to support scale-aware data discovery, thereby improving the alignment between user requirements and dataset granularity.

## 6. Conclusion

This study proposes a knowledge graph-driven multi-agent framework for intelligent geospatial data discovery. The framework combines a unified geospatial metadata ontology, knowledge graph construction, LLM-based intent parsing, graph retrieval, and answer synthesis into an interpretable end-to-end discovery process. Experimental results show that it improves intent matching accuracy, ranking performance, recall, and transparency compared with traditional discovery systems. More broadly, this study advances geospatial data discovery toward a more semantic, intent-aware, and intelligent paradigm, providing a practical foundation for next-generation intelligent and autonomous spatial data infrastructures. Future work will extend this framework from data discovery toward a broader autonomous pipeline that supports the generation of analysis-ready data for autonomous GIS.

## Disclosure statement

No potential conflict of interest was reported by the author(s).

## Data and codes availability statement

The source codes, graph database, evaluation data, and video demonstrations that support the findings of this study are available at https://figshare.com/s/1b5988fdad3b0239d9b1.

## Notes on contributors

Ruixiang Liu is a Ph.D. student in the Department of Geography at The Pennsylvania State

University. His research focuses on Autonomous GIS, geospatial knowledge graphs, large language models, and Agentic AI. He contributed to the idea, study design, methodology, implementation, and manuscript writing of this paper.

Zhenlong Li is an Associate Professor in the Department of Geography and Director of the Geoinformation and Big Data Research Lab at The Pennsylvania State University. His primary research field is GIScience with a focus on geospatial big data analytics, spatial computing, and geospatial AI with applications to disaster management, human mobility, and public health. He supervised the research and contributed to the idea, study design, methodology, and manuscript revision of this paper.

Ali Khosravi Kazazi is a Ph.D. student in the Department of Geography at The Pennsylvania State University. His research focuses on Autonomous GIS, large language models, and Agentic AI. He contributed to the implementation of this paper.

## Appendix A. Mapping of ontology

Table. A1. Mapping of geospatial metadata ontology and existing geospatial metadata standards.

| Entity | Geospatial Metadata Ontology | STAC Collection | FGDC CSDGM | CKAN |
|---|---|---|---|---|
| Dataset | id | id | / | id |
| | title | title | Identification_Information.Citation.Citation_Information.Title | title |
| | description | description | Identification_Information.Description | description |
| | url | links[rel='self'].href | / | / |
| Organization | id | / | / | organization.id |
| | title | providers.name | Identification_Information.Citation.Citation_Information.Originator | organization.title |
| | description | / | / | organization.description |
| Keyword | id | / | / | tags.id |
| | name | keywords | Identification_Information.Keywords | tags.name |
| License | id | / | / | license_id |
| | title | license | Identification_Information.Access_Constraints+ Identification_Information.Use_Constraints | license_title |
| Space | east | extent.spatial.bbox[2] | Identification_Information.Spatial_Domain. Bounding_Coordinates.East_Bounding_Coordinate | extras[key='bbox-east-long'].value |
| | west | extent.spatial.bbox[0] | Identification_Information.Spatial_Domain. Bounding_Coordinates. West_Bounding_Coordinate | extras[key='bbox-west-long'].value |
| | north | extent.spatial.bbox[3] | Identification_Information.Spatial_Domain. Bounding_Coordinates.North_Bounding_Coordinate | extras[key='bbox-north-lat'].value |
| | south | extent.spatial.bbox[1] | Identification_Information.Spatial_Domain. Bounding_Coordinates.South_Bounding_Coordinate | extras[key='bbox-south-lat'].value |
| Time | begin | extent.temporal.interval[0][0] | Identification_Information.Time_Period_of_Content.Range_of_Dates/Times.Beginning_Date | extras[key='temporal-extent-begin'].value |
| | end | extent.temporal.interval[0][1] | Identification_Information.Time_Period_of_Conte | extras[key='temporal-extent-end'].value |

|  |  |  | nt.Range_of_Dates/Times.Ending_Date |  |
|---|---|---|---|---|
| Resource | id | / | / | resources.id |
|  | title | links[rel='*'].rel | / | resources.title |
|  | description | / | / | resources.description |
|  | url | links[rel='*'].href | / | resources.url |
| Format | id | / | / | / |
|  | name | links[rel='*'].type | / | resources.format |

## Appendix B. Discovered datasets for representative use cases

Table. B1. Top 20 datasets generated by the graph retrieval agent for the climate data discovery.

| No. | Dataset | Topic | Space | Time | Total/Normalized Score | Synthesis Ranking by LLM |
|---|---|---|---|---|---|---|
| 1. | PRISM Daily Spatial Climate Dataset ANd (From Google Earth Engine Proxy for openEO) | daily air temperature Score: 0.8925 | [-125, 24, -66, 50] Score: 0.9702 | [19810101 00:00:00, 20251108 12:00:00] Score: 0.8174 | 0.6253/1.0000 | 1 |
| 2. | GRIDMET: University of Idaho Gridded Surface Meteorological Dataset (From Google Earth Engine Proxy for openEO) | daily air temperature Score: 0.8925 | [-124.9, 24.9, -66.8, 49.6] Score: 0.9833 | [19790101 00:00:00, 20251110 06:00:00] Score: 0.7963 | 0.6237/0.9975 | 2 |
| 3. | U.S. Daily Gridded Precipitation and Temperature Climate Normals for 2006-2020 (NCEI Accession 0259964) (From Data.gov) | daily temperature normals Score: 0.8906 | [-124.6875, 24.5625, -67.0208, 49.3542] Score: 0.9946 | [20060101 00:00:00, 20201231 00:00:00] Score: 0.6521 | 0.5965/0.9540 | 4 |
| 4. | NOAA nClimGrid-Daily Version 1 â€“ Daily gridded temperature and precipitation for the Contiguous United States since 1951 (From Data.gov) | maximum minimum temperature Score: 0.8488 | [-125, 24, -66, 50] Score: 0.9702 | [19510101 00:00:00, 20251231 00:00:00] Score: 0.5850 | 0.5657/0.9047 | 3 |
| 5. | PRISM Monthly Spatial Climate Dataset ANm (From Google Earth Engine Proxy for openEO) | monthly temperature Score: 0.9067 | [-125, 24, -66, 50] Score: 0.9702 | [18950101 00:00:00, 20251001 00:00:00] Score: 0.3833 | 0.5427/0.8679 | / |
| 6. | PRISM Monthly Spatial Climate Dataset AN81m [deprecated] (From Google Earth Engine Proxy for openEO) | monthly air temperature Score: 0.8454 | [-125, 24, -66, 50] Score: 0.9702 | [18950101 00:00:00, 20201201 00:00:00] Score: 0.3940 | 0.5265/0.8420 | / |

| 7. | A spatially comprehensive, meteorological data set for Mexico, the U.S., and southern Canada (NCEI Accession 0129374) (From Data.gov) | daily air temperature Score: 0.8925 | [-125.0000, 14.6563, -67.0000, 53.0000] Score: 0.7871 | [19500101 00:00:00, 20131231 00:00:00] Score: 0.5052 | 0.5262/0.8416 | 5 |
|---|---|---|---|---|---|---|
| 8. | U.S. Climate Divisional Dataset (Version Superseded) (From Data.gov) | maximum/minimum temperature Score: 0.8488 | [-125, 24, -66, 50] Score: 0.9702 | [18950101 00:00:00, 20251231 00:00:00] Score: 0.3827 | 0.5252/0.8400 | 10 |
| 9. | SST, Aqua MODIS, NPP, 0.0125Â°, West US, Day time (11 microns), 2002-present (1 Day Composite) (From Data.gov) | daily temperature composites Score: 0.8518 | [-155, 22, -105, 51] Score: 0.3415 | [20020624 12:00:00, 20220902 12:00:00] Score: 0.7237 | 0.4686/0.7494 | / |
| 10. | SST, NOAA POES AVHRR, LAC, 0.0125 degrees, West US, Day and Night, 2004-present (1 Day Composite) (From Data.gov) | daily temperature composites Score: 0.8518 | [-145, 22, -105, 51] Score: 0.3795 | [20040102 12:00:00, 20220902 12:00:00] Score: 0.6845 | 0.4683/0.7490 | / |
| 11. | SST, NOAA POES AVHRR, LAC, 0.0125 degrees, West US, Day and Night, 2004-present (1 Day Composite), degree_F (From Data.gov) | daily temperature composites Score: 0.8518 | [-145, 22, -105, 51] Score: 0.3795 | [20040102 12:00:00, 20220902 12:00:00] Score: 0.6845 | 0.4683/0.7490 | / |
| 12. | SST, NOAA POES AVHRR, LAC, 0.0125 degrees, West US, Day and Night, 2004-present (1 Day Composite), Lon+/-180 (From Data.gov) | daily temperature composites Score: 0.8518 | [-145, 22, -105, 51] Score: 0.3795 | [20040102 12:00:00, 20220902 12:00:00] Score: 0.6845 | 0.4683/0.7490 | / |
| 13. | SST, Aqua MODIS, NPP, East US, Daytime and Nighttime (11 microns), 2002-2011 (3 Day Composite) (From Data.gov) | day night temperature Score: 0.8609 | [-90, 20, -60, 50] Score: 0.4912 | [20020705 12:00:00, 20110911 12:00:00] Score: 0.4571 | 0.4479/0.7164 | / |
| 14. | ERA5 Daily Aggregates - Latest Climate Reanalysis Produced by ECMWF / Copernicus Climate Change Service (From Google Earth Engine Proxy for openEO) | daily air temperature Score: 0.8925 | [-180, -90, 180, 90] Score: 0.0436 | [19790102 00:00:00, 20200709 00:00:00] Score: 0.8417 | 0.4448/0.7114 | 6 |
| 15. | SST, Aqua MODIS, NPP, East US, Daytime and Nighttime (11 microns), 2002-2010 (8 Day Composite) (From Data.gov) | day night temperature Score: 0.8609 | [-90, 20, -60, 50] Score: 0.4912 | [20020705 12:00:00, 20101228 00:00:00] Score: 0.4297 | 0.4424/0.7076 | / |
| 16. | SST, GOES Imager, Day and Night, Western Hemisphere, 2000-2020 (1 Day Composite) (From Data.gov) | daily temperature composites Score: 0.8518 | [-179.975, -44.975, -30.025, 59.975] Score: 0.1682 | [20001201 12:00:00, 20200302 12:00:00] Score: 0.7662 | 0.4424/0.7075 | / |
| 17. | SST, GOES Imager, Day and Night, Western Hemisphere, 2000-2020 (1 Day Composite), Lon+/-180 (From Data.gov) | daily temperature composites Score: 0.8518 | [-179.975, -44.975, -30.025, 59.975] Score: 0.1682 | [20001201 12:00:00, 20200302 12:00:00] Score: 0.7662 | 0.4424/0.7075 | / |
| 18. | CHIRTS-daily: Climate Hazards Center InfraRed Temperature with Stations daily temperature data product (From Google Earth Engine Proxy for openEO) | daily maximum temperature Score: 0.8601 | [-180, -60, 180, 70] Score: 0.0599 | [19830101 00:00:00, 20161231 00:00:00] Score: 0.8307 | 0.4361/0.6975 | 8 |
| 19. | MOD11A1.006 Terra Land Surface Temperature and Emissivity Daily Global 1km [deprecated] (From Google Earth Engine Proxy for openEO) | daytime temperature Score: 0.8974 | [-180, -90, 180, 90] Score: 0.0436 | [20000224 00:00:00, 20221115 00:00:00] Score: 0.7763 | 0.4332/0.6928 | 9 |
| 20. | CPC Global Unified Temperature (From Google Earth Engine Proxy for openEO) | daily maximum temperature Score: 0.8601 | [-180, -90, 180, 90] Score: 0.0436 | [19790101 00:00:00, 20251110 00:00:00] Score: 0.7963 | 0.4260/0.6813 | 7 |

Table. B2. Top 10 datasets generated by the answer synthesis agent for the U.S. census data discovery.

| No. | Dataset | Score | Match Reason |
|---|---|---|---|
| 1. | TIGER/Line Shapefile, 2022, State, California, Primary and Secondary Roads | Raw Total Score: 0.7714<br>Normalized Score: 1.0000<br>Topic:0.9049, Format:1.0000, Organization: 0.8206, Space: 1.0000, Time: 0.5894 | • Exact geography match: California (state-level)<br>• Exact year match: TIGER/Line Shapefile 2022<br>• Direct topic match: Primary and Secondary Roads; Census Bureau TIGER/Line ZIP distribution |
| 2. | TIGER/Line Shapefile, 2022, Nation, U.S., Primary Roads | Raw Total Score: 0.5728<br>Normalized Score: 0.7426<br>Topic:0.9049, Format:1.0000, Organization: 0.8206, Space: 0.0070, Time: 0.5894 | • Direct topic match: Primary Roads (explicitly named)<br>• Correct year and format: 2022 TIGER/Line Shapefile, typically provided as ZIP<br>• Useful fallback if statewide CA layer is not sufficient (can subset to California) |
| 3. | TIGER/Line Shapefile, 2022, County, Inyo County, CA, All Roads | Raw Total Score: 0.5924<br>Normalized Score: 0.7679<br>Topic:0.9069, Format:1.0000, Organization: 0.8206, Space: 0.1020, Time: 0.5894 | • California coverage (Inyo County, CA) within the requested state<br>• Road network content (All Roads) can include primary roads segments<br>• Correct vintage/format: 2022 TIGER/Line Shapefile in ZIP |
| 4. | TIGER/Line Shapefile, 2022, County, Fresno County, CA, All Roads | Raw Total Score: 0.5888<br>Normalized Score: 0.7632<br>Topic:0.9069, Format:1.0000, Organization: 0.8206, Space: 0.0839, Time: 0.5894 | • California coverage (Fresno County, CA) relevant to the requested geography<br>• All Roads layer includes primary roads as a subset<br>• Matches 2022 TIGER/Line Shapefile ZIP delivery format |
| 5. | TIGER/Line Shapefile, 2022, County, Yolo County, CA, All Roads | Raw Total Score: 0.5700<br>Normalized Score: 0.7389<br>Topic:0.8925, Format:1.0000, Organization: 0.8206, Space: 0.0114, Time: 0.5894 | • California coverage (Yolo County, CA) relevant to the request<br>• All Roads provides road features that can be filtered to primary roads<br>• 2022 TIGER/Line Shapefile, commonly distributed as ZIP |
| 6. | TIGER/Line Shapefile, 2022, County, Yuba County, CA, All Roads | Raw Total Score: 0.5565<br>Normalized Score: 0.7214<br>Topic:0.8490, Format:1.0000, Organization: 0.8206, Space: 0.0092, Time: 0.5894 | • California coverage (Yuba County, CA) within requested state<br>• All Roads can be used to derive/select primary roads<br>• Correct year/format: 2022 TIGER/Line Shapefile ZIP |
| 7. | TIGER/Line Shapefile, 2022, State, Nevada, Primary and Secondary Roads | Raw Total Score: 0.6890<br>Normalized Score: 0.8932<br>Topic:0.9049, Format:1.0000, Organization: 0.8206, Space: 0.5879, Time: 0.5894 | • Same TIGER/Line 2022 road product type (Primary and Secondary Roads) from Census Bureau<br>• ZIP format match (TIGER/Line Shapefile distribution)<br>• Geography mismatch (Nevada) but structurally closest alternative road dataset |
| 8. | TIGER/Line Shapefile, 2022, State, Oregon, Primary and Secondary Roads | Raw Total Score: 0.5718<br>Normalized Score: 0.7413<br>Topic:0.9049, Format:1.0000, Organization: 0.8206, Space: 0.0021, Time: 0.5894 | • Same TIGER/Line 2022 road product type (Primary and Secondary Roads)<br>• Matches requested format (ZIP) and authoritative source (U.S. Census Bureau)<br>• Geography mismatch (Oregon) but relevant as comparable dataset |
| 9. | TIGER/Line Shapefile, 2022, State, Idaho, Primary and Secondary Roads | Raw Total Score: 0.5716<br>Normalized Score: 0.7410<br>Topic:0.9049, Format:1.0000, Organization: 0.8206, Space: 0.0009, Time: 0.5894 | • Same TIGER/Line 2022 road product type (Primary and Secondary Roads)<br>• ZIP format match and Census Bureau origin<br>• Geography mismatch (Idaho) but still relevant for the specific 'primary/secondary roads' theme |
| 10. | TIGER/Line Shapefile, 2022, State, Arizona, Primary and Secondary Roads | Raw Total Score: 0.5667<br>Normalized Score: 0.7347<br>Topic:0.8581, Format:1.0000, Organization: 0.8206, Space: 0.0469, Time: 0.5894 | • Same TIGER/Line 2022 road product type (Primary and Secondary Roads)<br>• ZIP format match and produced by the U.S. Census Bureau<br>• Geography mismatch (Arizona) but closer topical match than non-road California layers |

Table. B3. Health data discovery results (showing Top 3).

| **Question** | **Dataset** | **Match Reason** |
|---|---|---|
| I am looking for HIV data in District of Columbia. | HIV and AIDS Clinic | • Direct topic match: HIV/AIDS (clinic-focused dataset)<br>• Most relevant to finding HIV-related services/providers in DC context<br>• High topic score (0.94) among candidates |
| | AIDS Drug Assistance Program | • Direct topic match: AIDS Drug Assistance Program (HIV/AIDS support program data)<br>• Useful for HIV program coverage/access information relevant to DC residents<br>• High topic score (0.94) among candidates |
| | DC Health Planning Neighborhoods to Census Tracts | • Strong DC geographic relevance (Space=1.00) for District of Columbia analyses<br>• Provides neighborhood-to-census-tract crosswalk to map/aggregate HIV indicators<br>• Supports joining HIV metrics to DC health planning geographies |
| Please find opioid use datasets published by the City of Tempe, licensed under the Creative Commons Attribution license, and provided in GeoJSON format. | Opioid EMS Calls | • Direct opioid-related dataset (EMS calls) matching the 'opioid use' topic<br>• Matches key constraints: City of Tempe organization and CC BY license<br>• GeoJSON format match (high format score) |
| | Biomarker Opioids (open data) | • Strong opioid relevance (opioids biomarker data) aligned with opioid use monitoring<br>• Published by City of Tempe and licensed under Creative Commons Attribution<br>• Available in GeoJSON per candidate format score |
| | 1.31 Opioid EMS Calls (summary) | • Opioid-specific EMS calls summary relevant to opioid use/overdose activity<br>• City of Tempe publisher and CC BY license match the request<br>• GeoJSON format match (high format score) |